\documentclass[lettersize,journal]{IEEEtran}
\usepackage[subpreambles=true]{standalone}
\usepackage{import}
\usepackage[subpreambles=true]{standalone}

\usepackage{amsmath} % assumes amsmath package installed
\usepackage{graphicx}
\usepackage{makecell}
\usepackage{amsfonts}
\usepackage{amssymb,bm}
\usepackage{import}
\usepackage{xifthen}
\usepackage{pdfpages}
\usepackage{transparent}
\usepackage{subcaption}

\usepackage[ruled,vlined,linesnumbered]{algorithm2e}
\usepackage{algpseudocode}
\usepackage{float, url}
\usepackage{multirow}
\usepackage{booktabs}

 \usepackage{soul}

\newcommand \figref[1]{Fig. \ref{#1}}

\newcommand\StateSpace[1]{\mathbf{\mathcal #1}}
\newcommand\NaturalNumbers{\mathbb{N}}
\newcommand\Probability{\mathbb{P}}
\newcommand\MinusSet{\backslash}
\newcommand\Real{\mathbb{R}}
\renewcommand\vector[1]{\bm{#1}}
\newcommand\ExpectedValue{\mathbb{E}}
\newcommand\BehavesLike{\sim}
\newcommand\Norm[1]{\|#1 \|_2}
\newcommand\Gradient{\nabla}
\newcommand\Normal{\mathcal{N}}

\usepackage{tikz, pgfplots}
  \pgfplotsset{compat=newest}
  \usepgfplotslibrary{external}
\title{\LARGE \bf
Improving Path Planning Performance through Multimodal Generative Models with Local Critics
}

\author{Jorge Ocampo Jimenez and Wael Suleiman%
\thanks{Authors are with Electrical and Computer Engineering Department, Universit\'e de Sherbrooke, Quebec, Canada  (e-mail: \{Jorge.Ocampo-Jimenez; Wael.Suleiman\}@USherbrooke.ca)}%

}

\begin{document}
\maketitle
\thispagestyle{empty}
\pagestyle{empty}

\begin{abstract}
This paper presents a novel method for accelerating path planning tasks in unknown scenes with obstacles by utilizing Wasserstein Generative Adversarial Networks (WGANs) with Gradient Penalty (GP) to approximate the distribution of the free conditioned configuration space. Our proposed approach involves conditioning the WGAN-GP with a Variational Auto-Encoder in a continuous latent space to handle multimodal datasets. However, training a Variational Auto-Encoder with WGAN-GP can be challenging for image-to-configuration-space problems, as the Kullback-Leibler loss function often converges to a random distribution. To overcome this issue, we simplify the configuration space as a set of Gaussian distributions and divide the dataset into several local models. This enables us to not only learn the model but also speed up its convergence. We evaluate the reconstructed configuration space using the homology rank of manifolds for datasets with the geometry score. Furthermore, we propose a novel transformation of the robot's configuration space that enables us to measure how well collision-free regions are reconstructed, which could be used with other rank of homology metrics. Our experiments show promising results for accelerating path planning tasks in unknown scenes while generating quasi-optimal paths with our WGAN-GP. The source code is openly available\footnote{\url{https://bitbucket.org/joro3001/multiwgangp/}}.
\end{abstract}


\section{Introduction}

Random sampling algorithms are a popular choice for finding a collision-free path through a robot's configuration space (CS) from an initial state to a goal region. However, these algorithms can be time-consuming, as they involve exploring the configuration space randomly. This becomes even more challenging when the CS is complex and difficult to model analytically, which is often the case in practical applications. Therefore, there is a need to develop more efficient and accurate methods for path planning, especially in scenarios where random sampling algorithms may not be suitable.

\begin{figure}[tb]
\centering
\begin{subfigure}{0.45\columnwidth}
	\def\svgwidth{\textwidth}
    %
    %\import{#1.tex}
    %\def\svgwidth{\columnwidth}
    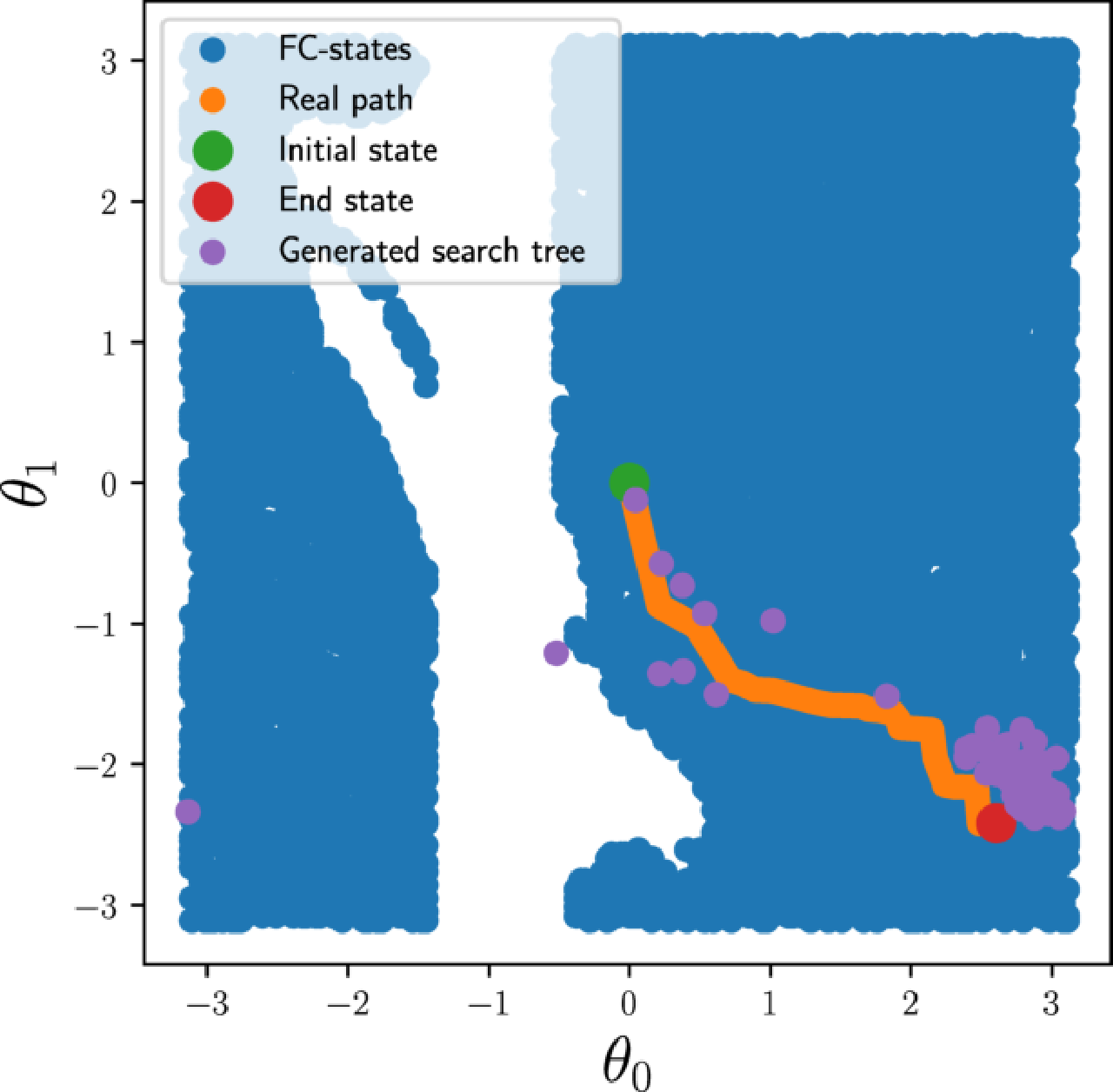

    \caption{Approximation of the free-search tree of a path with WGAN-GP and a VAE for an unseen scenario}

    \label{fig:simpleExtrapolationPath}
\end{subfigure}
\hfill
\begin{subfigure}{0.45\columnwidth}
	\def\svgwidth{\textwidth}
    %
    %\import{#1.tex}
    %\def\svgwidth{\columnwidth}
    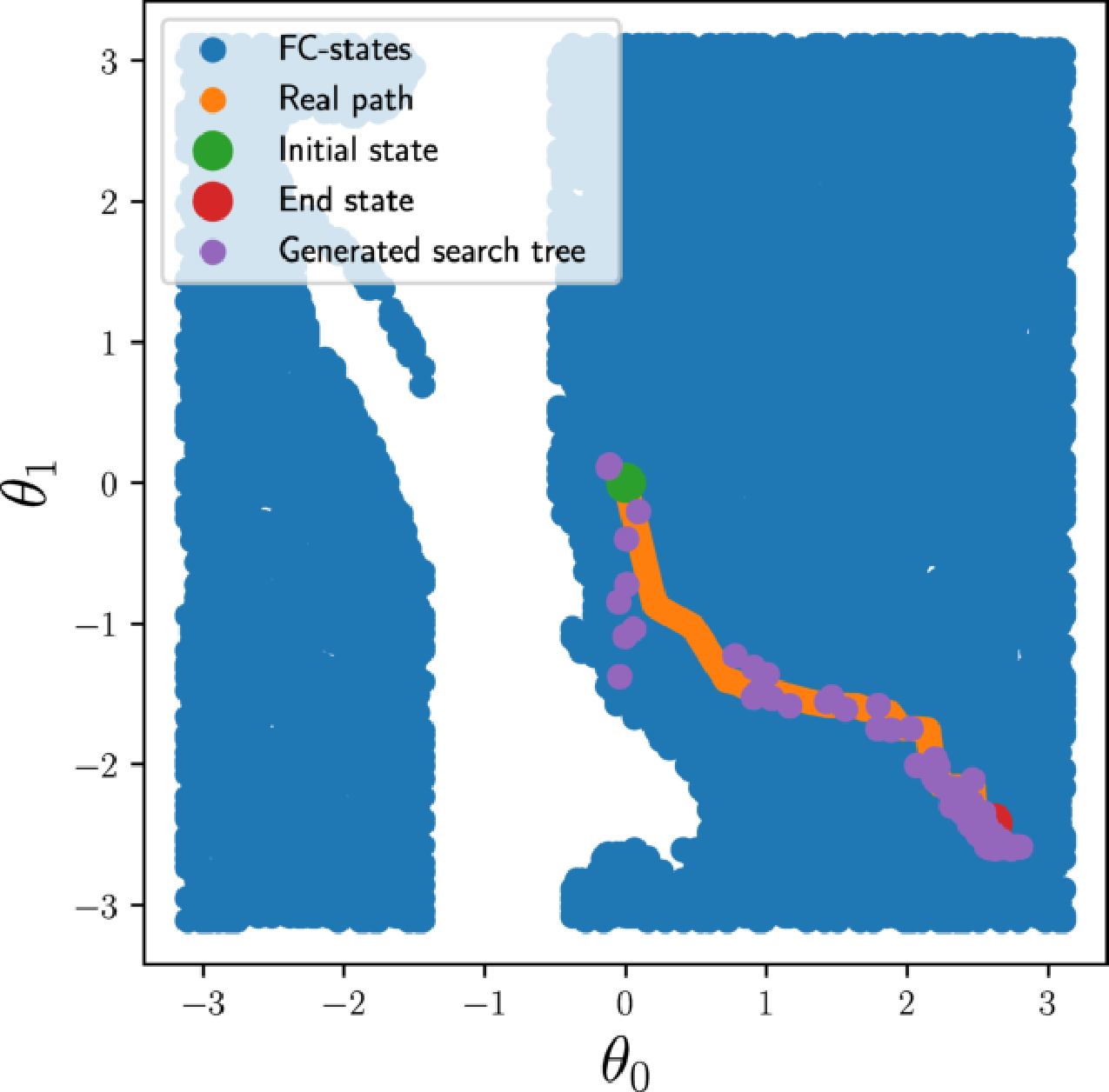

    \caption{Approximation of the free-search tree of a path with MultiWGAN-GP for an unseen scenario space}

    \label{fig:extrapolationOneClusterPath}
\end{subfigure}
\caption{Search tree for a path planning task of unseen scenarios.}
\label{fig:comparisonPaths}
\end{figure}

Machine learning techniques, such as neural network generative models like Variational Auto-Encoder (VAE) \cite{qureshi2019motion} and Generative Adversarial Networks (GANs) \cite{DBLP:journals/ral/LembonoPJC21}, have been applied to improve the efficiency of random sampling algorithms and bias the distribution of samples towards collision free (CF)-states. While previous works have successfully used GANs \cite{DBLP:journals/ral/LembonoPJC21}, \cite{JAS-2021-0110}, \cite{Li2021EfficientHG} to generate inverse and forward kinematics, there has been little research on using GANs to model multi-modal distributions, such as RGB images encoded as latent vectors to CF states. Therefore, this paper aims to address this gap by investigating the use of GANs for this purpose. To overcome challenges such as unstable training processes and the potential for unexpected results when the conditioning input has not been seen before, this work proposes encoding latent vectors with a VAE. Furthermore, this paper aims to evaluate the general ability of GANs to generate new samples for new scenarios as shown in \figref{fig:comparisonPaths}. By investigating the use of GANs in this way, this research could lead to significant improvements in the efficiency and accuracy of path planning algorithms in a variety of practical applications.

This work utilizes an RGB image of the robot's image-scenario to encode the input condition. The aim is to convert discrete training data into a continuous representation, which empowers the generator to create unconstrained configuration states from input data that was not included in the training phase. This issue is akin to interpolating between two recognized positions of obstacles in the image-scenario. In a similar vein to \cite{park2019SPADE}, which addresses image-to-image translation, an encoder is leveraged to embed an image into a random vector to capture the style of an image-to-image GAN's mask. Our work differs from this as the encoder is utilized to influence the latent vector using an RGB input of the obstacles, directing the GAN towards areas where the current model's parameters are more prone to generating valid configurations.

\begin{figure*}
\begin{center}
\includegraphics[width=\textwidth]{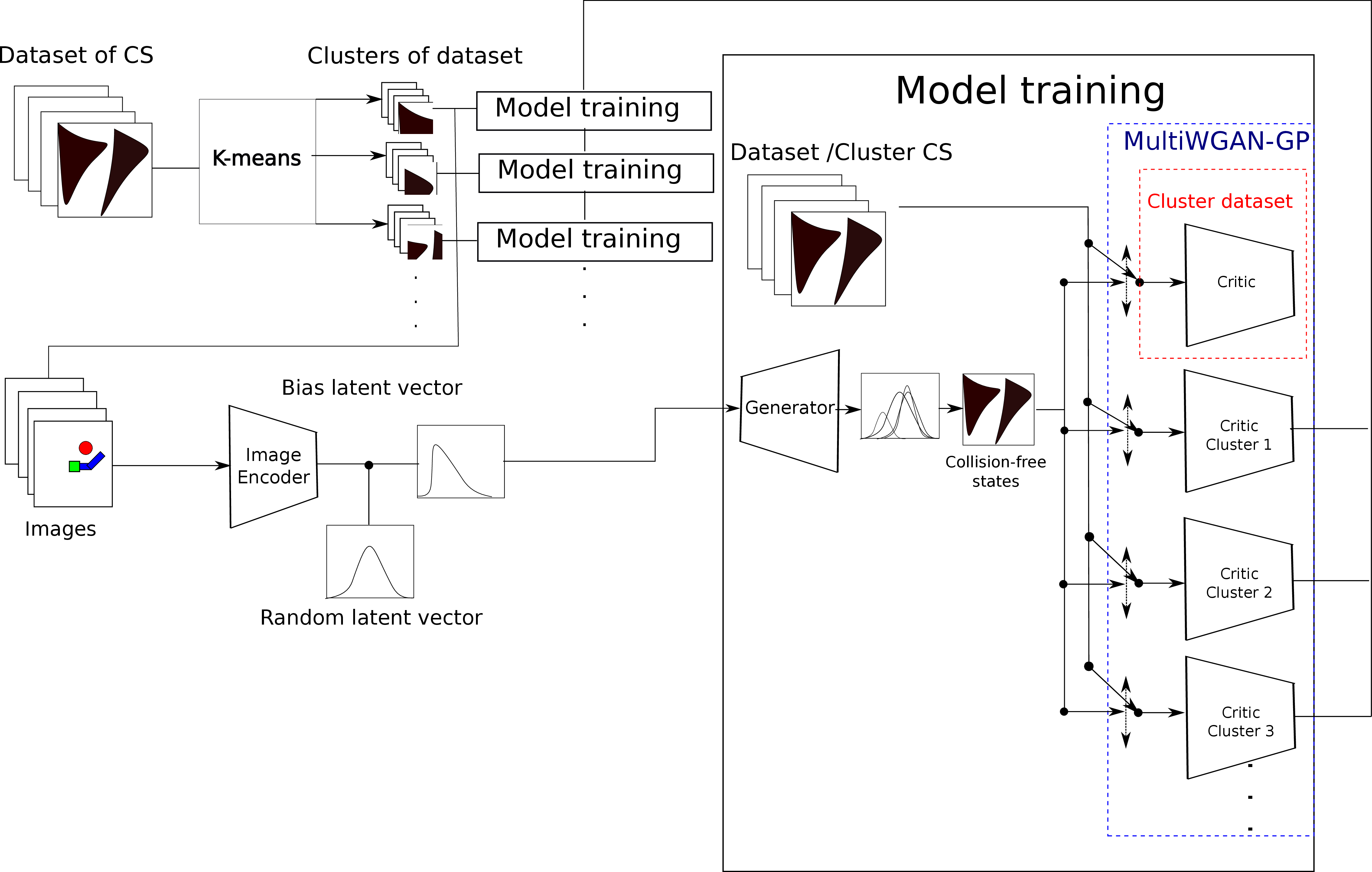}
\caption{Our proposed model generates the Gaussians of different regions of the CS of the robot. RGB images of the obstacles are used as input to bias the latent vector to the correct region of the CS.}
\label{fig:model}
\end{center}

\end{figure*}

Our model was trained using the forward kinematics from the simulations presented in \cite{DBLP:journals/corr/abs-1802-05637}. We created multiple scenarios by placing random obstacles around a 2 DOF manipulator and utilized the robot's obstacles represented as images to train our model to reconstruct its configuration space. To avoid defining a fixed clipping interval, we employed a Wasserstein loss function \cite{arjovsky2017wasserstein} with gradient penalty \cite{10.5555/3295222.3295327} for training the model. Our experiments have shown that dividing the configuration space into subsets enhances the convergence of the algorithm and simplifies the training process. This is achieved by reducing the problem to determining the translations of the means of a set of Gaussians, which facilitates the search and prevents model collapse. The architecture for our method is illustrated in \figref{fig:model}. Notably, our approach outperformed standard training methods and significantly enhanced the reconstruction of the configuration space.

We evaluated the performance of our model using the Geometry Score (GS) metric \cite{conf/icml/KhrulkovO18}. This allowed us to gauge the impact of proposed model on the reconstruction of the collision-free CS by comparing the rank of the homology of the manifolds. To establish a baseline for comparison, we employed a WGAN-GP conditioned by a VAE that generated CF-states directly. Our experiments revealed that our model converges faster and more accurately to the original CS. We anticipate that our model holds the potential to be replicated and enhanced for tasks that necessitate the generation of multi-modal CSs for motion planning problems.

\section{Original Contributions}

This paper offers several novel contributions, including:
\begin{enumerate}

    \item Demonstrating the possibility of training WGAN-GP for path planning tasks and image-to-CS models for CS reconstruction.
    \item Introducing a new architecture that improves the training of WGAN-GP when conditioned by an Auto-Encoder.
    \item Presenting a new method to generate configuration spaces that works effectively for scenarios not previously encountered.
    \item Extending the rank of homology scores to quantify the accuracy of a reconstructed CS compared to the original data.
    \item In our approach, we propose perturbing the encoder to handle situations where the generator fails to produce collision-free states without relying on sampling from alternative distributions.

\end{enumerate}
Compared to other similar works where the planner's sampling is biased, our method offers several advantages, for instance:
\begin{itemize}
\item By using both GANs and VAEs, our model can handle noisy and non-previously trained scenarios while generating quality WGAN samples without requiring the identification of new clipping intervals for training. 
\item Our image-to-CS model has the potential to be extended to reconstruct higher-dimensional CSs for robotic tasks.
    \item  Unlike traditional methods that rely on finding a suitable ratio between a uniform or biased distribution for RRT-based planners, our approach focuses on manipulating the encoder parameter to enhance the generation of collision-free states.
\end{itemize}
 
\section{Related Work}

The utilization of learning by demonstration has proven to be an effective approach in multiple studies aimed at enhancing the performance of sampling-based random planners \cite{9712347}. A common technique employed is the use of an autogenerative model that learns a map linking the robot's configuration space and the image-scenario with a reduced number of samples from the complete distribution. In recent years, deep neural models have gained significant popularity in this field due to their ability to handle vast quantities of input data, such as image or cloud point representations of the robot's scenario, as well as an extensive range of examples, including the robot's possible configurations and the number and location of obstacles present in the workspace.

Generative models are widely utilized in the context of Rapidly-exploring Random Trees (RRT)-based algorithms \cite{Kuffner} for two primary purposes: to provide a bias to the sampler or to act as a heuristic for the cost function. The model guides the algorithm towards lower cost paths by considering the condition of the scenario.

The application of neural networks for learning the sampling distributions of biased sampling-based planners was first introduced in \cite{LearningSampling}. The study utilized a conditional variational autoencoder to identify areas in the state space that held promise based on the initial and goal states, as well as the obstacles present in the scenario. This enabled the sampler of RRT-based algorithms to be biased, resulting in more efficient path planning.

In another study, \cite{qureshi2019motion}, an encoder was also utilized to capture the environment's information, and the sampler was conditioned on raw sensor data or voxelized output embedded in the latent space. The encoded information was then utilized by a planning network, in combination with the current and goal states, to generate the next state. This model is capable of biasing the sampler of RRT* \cite{Karaman} and has been tested on high dimensional configuration spaces.

Although VAE methods have proven effective in reducing the computation time of sampling, GANs have demonstrated superior results in tasks such as image-to-image generation and dataset generation, which could potentially improve the success rate of planned paths.

The study presented in \cite{teachingRobot} employs inverse reinforcement learning to determine the weights of the RRT*'s cost function, based on the expected behavior of a robot in environments previously occupied by humans. This approach guides the planner towards the desired path. However, it may not be suitable for dynamic environments where the weights cannot be modified without compromising the asymptotic optimality of the algorithm.

The research presented in \cite{JAS-2021-0110} utilizes 2D working spaces as inputs for a conditional GAN. The GAN is conditioned on the RGB representation of the initial and final points of the path, as well as the map of the working space. The generator is trained using two discriminators, with one responsible for the obstacle map and the other for the initial and final goal states represented in the working space. The resulting algorithm has an impressive success rate of approximately 90\% for generating connected configurations.

The work in \cite{Li2021EfficientHG} presents an approach where GANs are utilized to bias a RRT-based planner by incorporating Encoders and Decoders directly as hidden layers in the generator. The initial state, map, and latent vector are given as input to the encoder's input layer, while the decoders output a 2D representation of the path. The generator is then able to output the path as an output image. The method treats the path as an image-to-image model. While it does not provide information on running times, it is reported that the algorithm takes fewer iterations to achieve a lower cost than RRT*.

Although GANs have been successful in generating feasible configurations through 2D representations, where the workspace and configuration space overlap, their application in high-dimensional configuration spaces or in models where the workspace is different from the configuration space is still unclear. The current representation limits existing techniques to only image-to-image generation, thus reducing their applicability. Therefore, there is a need to develop new methods that can overcome these limitations and expand the application of GANs to a wider range of scenarios, including high-dimensional configuration spaces and non-overlapping workspace and configuration spaces.

The research presented in \cite{DBLP:journals/ral/LembonoPJC21} explores the use of a GAN to learn the inverse kinematics of high-dimensional robots. The model is conditioned on the target working space position of the end-effectors, enabling the generation of samples in high dimensional configuration spaces, which was previously not feasible. It is important to note, however, that the conditioning is not directly based on sensor data or the current state of the scenario.

Although new advances in generative neural network models have been proposed, such as the diffusion generation approach in path planning \cite{huang2023diffusion}, their use for random sampling-based planners is limited to cases where there are time constraints to obtain a collision-free path. One common issue with diffusion-based methods is speed, and sampling the diffusion generator multiple times can be very costly.

Our approach builds upon previous work and overcomes some of their limitations. We use a multimodal generator, similar to \cite{qureshi2019motion}, but train it using a WGAN-GP \cite{10.5555/3295222.3295327} to generate collision-free configurations. This improves the quality of the generated configuration states and enables quasi-asymptotically-optimal planning in high dimensional configuration spaces with time constraints.

In contrast to \cite{10.1109/ICRA48506.2021.9561472}, we use a VAE to encode the conditioning from the obstacles image-scenario of the WGAN-GP generator, enabling us to generate configuration states that closely match those encountered in previously seen scenarios and explore new configurations. We also use the gradient penalty technique, which increases the capacity of the discriminator to learn complex features without requiring tuning of clipping parameters.

Our approach can be extended to higher dimensional configuration spaces, making it a versatile and effective method for sampling-based path planning and CS-reconstruction. In the case of path planning, we can embed the problem in higher-dimensional Euclidean spaces using the same model architecture.

\section{Problem Formulation}

The objective of this research is to develop a method for approximating the CS of a robot by leveraging information from its obstacles represented as an image-scenario, with the ultimate goal of enhancing the performance of a path planner. This is accomplished by training a model to learn the mapping from the image-scenario to the CS, allowing for more efficient sampling and speeding up the planning process. The proposed method has the potential to significantly enhance the performance of robotic systems by reducing the computational cost of planning, while still producing optimal paths.

Mathematically speaking, a path planning problem is defined by a state space
$\StateSpace{X}$$=[0,1]^d$
with dimension
$d \in \NaturalNumbers, d\geq 2$. $\StateSpace{X}_{obs}$
is defined as the set of obstacles that corresponds to the collision sates, it also defines the free state space $\StateSpace{X}_{free}=\StateSpace{X} \MinusSet \StateSpace{X}_{obs}$, with initial state
$\vector{x}_{0} \in \StateSpace{X}_{free}$
and a  set of goal states
$\StateSpace{X}_{goal} \subset \StateSpace{X}_{free}$
. A path is a continuous function
$s:[0,1]\rightarrow \Real^d$
, and it is collision-free if
$s(\tau)\in \StateSpace{X}_{free}$
for all
$\tau \in [0,1]$
and feasible when it is collision-free and
$s(0)=\vector{x}_{0}$ and $s(1) \in \StateSpace{X}_{goal}$.

Finding a feasible path in the CS of a robot is known to be PSPACE-complete \cite{10.5555/1213331}, which means that it is computationally intractable for most practical applications. As a result, researchers have developed sampling-based motion planning algorithms as a means of finding paths in high-dimensional CSs. These algorithms operate by randomly sampling configurations and connecting them to form a path to the goal states. Completeness, or finding a solution if one exists, requires drawing a sufficient number of uniformly distributed random samples. Asymptotic optimality, where the path cost converges to the optimal solution, can be achieved by systematically connecting the nodes of the search tree \cite{LearningSampling}.

To improve the efficiency of sampling-based motion planners, researchers have proposed various methods to bias the path towards the goal. One such method is to learn a probability distribution over the free states ($\StateSpace{X}_{free}$) based on the robot's scenario, which can guide the sampling process to explore regions of the CS that are more likely to lead to the goal.
As a result, the algorithm can decrease the time spent exploring regions of the CS that are not useful for finding a path. This speeds up the planning process and has the potential to substantially enhance the efficiency of sampling-based motion planning algorithms, making them more practical for real-world applications.

\section{Methodology}

We propose a novel approach for speeding up sampling-based motion planning algorithms by generating $\StateSpace{X}_{free}$ states with additional properties such as feasibility and connectivity with the current path. Our approach uses a GAN to sample from a learned distribution over $\StateSpace{X}_{free}$, which biases the sampling process towards regions of the CS more likely to lead to CF-states. Specifically, we use a WGAN-GP to generate high-quality collision-free configurations without the need for finding a suitable clipping interval. This replaces the uniform distribution usually used for sampling $\StateSpace{X}_{free}$ and leads to faster query times.

\subsection{Generative Model}
GAN are mainly defined by the loss function in Eq. \eqref{eq:1}:

\begin{equation}
\begin{split}\label{eq:1}
\min_G \max_D L(D,G)=\ExpectedValue_{x\BehavesLike p_r(\vector{x})}[\log D(\vector{x})] + \\
\ExpectedValue_{z\BehavesLike p_z(\vector{x})}[\log(1-D(G(\vector{z})))]
\end{split}
\end{equation}
where $p_r$ and $p_z$ represent respectively the distributions over the multidimensional real data $\vector{x}$ and the noise input vector $\vector{z}$. $D$ is a binary classifier that distinguishes between $p_r$ and the distribution of the generator $G$, $p_g$. The loss function represents a minimax game in which $G$ tries to deceive $D$ to classify $p_g$ as $p_r$. If Nash equilibrium is reached, $G$ will be able to generate new samples that closely resemble to $p_r$ each time that an input noise $z$ is given to $G$ \cite{10.5555/2969033.2969125}. However, directly using the proposed model to bias the sampling process poses a challenge, as it does not allow for conditioning the sampler on information about the current state of the robot or the scenario.

To address this, \cite{10.5555/2969033.2969125} proposes a conditional GAN that permits conditioning on known information about the distribution. In the context of path planning, this extra information can relate to the current state of the robot in various spaces, such as the obstacles in a specific scenario. The central idea is to concatenate a portion of the known information from $p_r$ in $G$ and $D$ in Eq. \eqref{eq:cgan} \cite{mirza2014conditional}:

\begin{equation}
\begin{split}\label{eq:cgan}
\min_G \max_D L(D,G)=\ExpectedValue_{x\BehavesLike p_r(\vector{x})}[\log D(\vector{x}|\vector{y})] + \\ \ExpectedValue_{z\BehavesLike p_z(\vector{x})}[\log(1-D(G(\vector{z}|\vector{y})))]
\end{split}
\end{equation}

where $\vector{y}$ can represent any form of auxiliary information. However, obtaining Nash equilibrium becomes challenging when the two loss models are updated independently \cite{10.5555/3157096.3157346}. If $p_g$ and $p_r$ are situated in low-dimensional manifolds, they will almost surely be disjoint \cite{Arjovsky2017TowardsPM}. Good discriminators can cause the gradient to vanish, which means that $G$ will only generate a small number of samples. 

In order to improve the training stability of GAN models, a method was proposed in \cite{arjovsky2017wasserstein} which employs the Earth-Mover distance to measure the similarity between $p_g$ and $p_r$. This approach offers the benefit of providing smooth measures even in scenarios where the distributions are completely overlapping or disjoint. The modified version of this method, which utilizes Kantorovich-Rubinstein Duality, is expressed as follows:

\begin{equation}
W(p_r,p_g) = \max_{w \in  W} \ExpectedValue_{x \BehavesLike p_r}[f_w(\vector{x})]-\ExpectedValue_{z\BehavesLike p_r(z)}[f_w(g_\rho (\vector{z}))]
    \label{eq:wgan}
\end{equation}
where $f_w$ is a K-Lipschitz continous function parameterized by $w$ in a compact parameter space $W$, and the weights of the network $g$ are $\rho$. Initially, in \cite{arjovsky2017wasserstein}, weight clipping was proposed as a method to stabilize the training of a WGAN model. However, choosing the right clipping parameters can be challenging, and setting them to values that are too large or too small can result in slow convergence or vanishing gradients. To address this issue, \cite{10.5555/3295222.3295327} introduced a gradient penalty approach that penalizes the model gradients if the Lipschitz constraint is violated. Specifically, if $f$ has a gradient norm greater than 1, a penalty term is added to the loss function to encourage the model to stay within the Lipschitz constraint as follows:

\begin{equation}
    \begin{split}
        L=
\underbrace{
 \ExpectedValue_{x \BehavesLike p_r} [f_w(\vector{x})] - \ExpectedValue_{z\BehavesLike p_r(z)} [f_w(g_\rho (\vector{z}))]
}_{\text{Original critic loss}} \\
+
\underbrace{
\lambda \ExpectedValue_{\hat{x}\BehavesLike p_{\hat{x}}}
[
 (
\Norm{\Gradient_{\hat{x}}f_w(\vector{\hat{x}})}
  -1
 )^2
]
}_{\text{Gradient penalty}}
    \end{split}
    \label{eq:wgangp}
\end{equation}
where $\lambda$ is a penalty coefficient to weight the gradient penalty, $\vector{\hat{x}}$ sampled from $g_{\rho}$ and $\vector{x}$ with $t$ uniformly sampled between $0$ and $1$:
\begin{equation}
\vector{\hat{x}}=t\vector{x}+(1-t)g_\rho(\vector{z})\text{ with }0\leq t \leq 1
    \label{eq:gangp}
\end{equation}
\subsection{Image conditioning}

Instead of conditioning the GAN directly on the input image, we used a VAE \cite{Kingma2014AutoEncodingVB}.  This autoencoder is utilized to map the input image of the current scenario into a latent vector $\vector{z}$. This latent vector is then used as the input to the WGAN-GP in \eqref{eq:gangp}. By transforming the image space into a parametrized multivariate normal distribution with vector mean $\mathbf{0}$ and vector standard deviation $\mathbf{1}$, the autoencoder aims to represent the image as closely as possible to the Gaussian distribution while preserving enough information to reconstruct the image using the latent vector $\vector{z}$. This enables us to generate new obstacle configurations that interpolate between data points from the training data during inference. This transformation is particularly valuable for path planning because it ensures that when the model receives a new scenario input, it generates an interpolation of the cost space based on the examples it was trained on.

To overlap the different samples from the training model, we included the Kullback-Leibler divergence loss function during the training of the WGAN-GP. The divergence loss is defined as the Kullback-Leibler divergence between two probability density functions $p(\cdot)$ and $q(\cdot)$ and is given by:
\begin{equation}
D_{KL}(p||q)=\int_{-\infty}^\infty p(x)\log \left( \frac{p(x)}{q(x)} \right)dx
\end{equation}
If the images are real valued, we use $\Normal(\mu_i,\sigma^2_i)$ as the likelihood and a $\Normal(\mathbf{0},\mathbf{1})$ as the prior, and the term simplifies as shown in \cite{Cinelli2021}: 

\begin{equation}
D_{KL}=\sum^n_{i=1}\sigma^2_i+\mu^2_i-\log (\sigma_i)-1
    \label{eq:kl}
\end{equation}

To train the encoder, we added Eq. \eqref{eq:kl} alongside the generation loss from \eqref{eq:wgan} to get:
\begin{equation}
    \min_G \left\{ -\ExpectedValue_{z \BehavesLike p_r(z)}[f_w(g_\rho (\vector{z}))]+ D_{KL}(q(\vector{z}|\vector{x})||p_r(\vector{z})) \right\}
    \label{eq:generatorAndVAE}
\end{equation}
 where $q(\vector{z}|\vector{x})$ is the latent distribution given the input image.

\subsection{Evaluating the reconstruction of CS}

An essential requirement for accurately reconstructing the CS in path planning problems is the ability to differentiate between CF states and states in collision. Collision states can be visualized as voids or holes in the CS of the robot. To evaluate the quality of the reconstructed distribution of collision-free states, we have chosen to use the Geometriy Score (GS) \cite{conf/icml/KhrulkovO18} as a metric. This score effectively measures how well the reconstructed CS represents the CF space and identifies the collision states.

The GS is a metric that utilizes the topological properties of a manifold, specifically the homology group formed by the quotient space between cycles and boundaries of graphs of discrete elements of the CS. Essentially, the GS measures the number of holes present in the CS and how well they are identified by the reconstructed space, providing a reliable assessment of the reconstruction's quality.

To define a hole in the context of a topological set, we create subsets of the space known as $k$-simplexes. The elements of these subsets can be interpreted as vertices, and the subsets spanned by the $k$-simplexes as faces. The homology group can then describe the holes, with the kernel of the cycles and the image of the boundaries numerically defining the quantity of the $k^{th}$ dimensional holes in the topology formed by the edges constructed by the $k^{th}$ simplexes.

Using the GS, we form $k$-simplices by selecting a neighborhood of random points from the reconstructed CSs as landmarks $\mathbb{L}$ from the dataset $\vector{X}$. These landmarks are then connected to the closest vertices using a changing parameter $\epsilon$, and the resulting $k$-simplices and faces are used to compute the rank of the homology (Betti number).

By varying the value of $\epsilon$, we can generate different $k$-simplices and different ranks for each $\epsilon-$homology. To estimate the score, we identify topological features that persist over a significant range of $\epsilon$. This measurement is known as the Mean Relative Living Times (MRLT), which comprises the Relative Living Times for each possible number of holes $t$, as calculated using the following equation:

\begin{equation}
    \label{eqn:RLT}
\text{RLT}(t,k,X,\mathbb{L})\triangleq \frac{\mu \{ \alpha \in [0, \alpha_{max}]: \beta_k (\alpha )=t \}}{ \alpha_{max}}
\end{equation}
where $\mu$ represents the mean, $\alpha$ is a relaxation parameter that generates a sequence of simplicial complexes, the maximum value of $\alpha$, denoted by $\alpha_{max}$, can be defined as a linear function of the distance between the cloud points, and $t$ is a positive integer number. $\beta_k$ is the $k^{th}$ Betti number, which represents the rank of the homology group of dimension $k$, it is defined by:

\begin{equation}
\beta_k (\alpha )\triangleq |\{ [b_t,d_t]\in  \mathcal{I}_k:\alpha \in [b_t,d_t]\} |
\end{equation}
$\mathcal{I}_k=\{[b_t,d_t] \}^s_{t=1}$ is a collection of $s \in \NaturalNumbers$ persistence intervals in a fixed $k$ dimension.

We can measure the topological activity associated with each value of $\beta(\alpha)$ by utilizing the RTL defined in \eqref{eqn:RLT}. This metric can be computed using the dataset $\vector{X}$ and a random selection of data points/landmarks $\mathbb{L}$ for a given number of holes $t$.

To obtain a comprehensive evaluation of the model, the MRTL is used to estimate the expected value across various numbers of holes:
\begin{equation}
\text{MRLT}(t,k,\vector{X})\triangleq \ExpectedValue_\mathbb{L}[RLT(t,k,\vector{X},\mathbb{L})]
\end{equation}
wich defines the the certainty about the number of k-dimensional holes on average.

To compare two datasets, $\vector{X}_1$ and $\vector{X}_2$, the GS is defined as the expected square difference of the datasets MRLT:
\begin{equation}
\resizebox{\columnwidth}{!}
     {%

    $\text{GS}(\vector{X}_1,\vector{X}_2)\triangleq \sum^{t_{max}-1}_{t=0}\left(\text{MRLT}(t,1,\vector{X}_1)
    -\text{MRLT}(t,1,\vector{X}_2)\right)^2$
     }
    \label{eqn:GeomScore}
\end{equation}
with $i_{max}$ being an upper bound on $\beta_k(\alpha)$.

\begin{figure}[ht]
\centering
\begin{subfigure}{0.45\columnwidth}
	\def\svgwidth{\textwidth}
    %
    %\import{#1.tex}
    %\def\svgwidth{\columnwidth}
    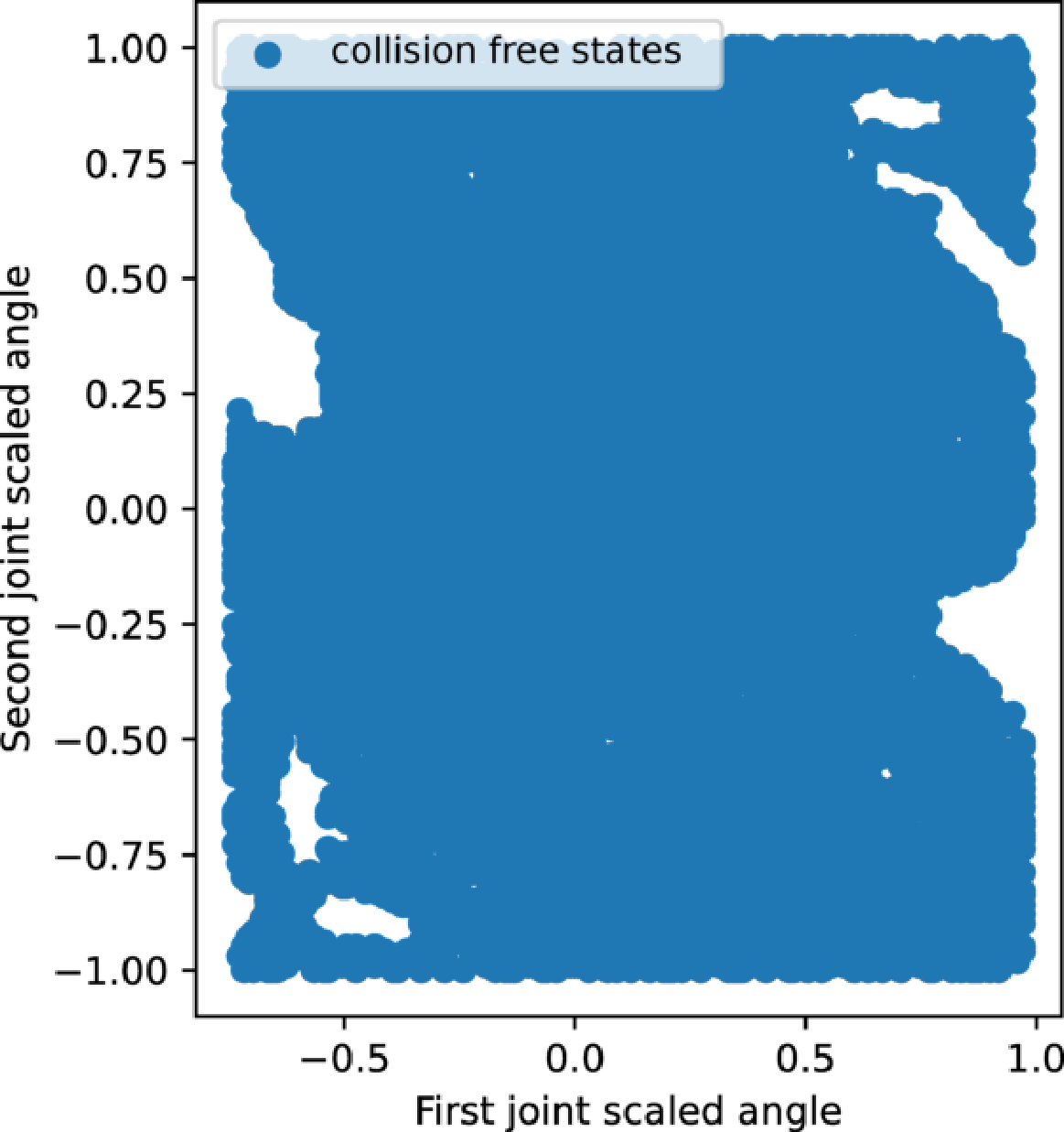

    \caption{CS where the obstacle constrains second link of the manipulator}

    \label{fig:CS0comparisonGS}
\end{subfigure}
\hfill
\begin{subfigure}{0.48\columnwidth}
	\def\svgwidth{\textwidth}
    %
    %\import{#1.tex}
    %\def\svgwidth{\columnwidth}
    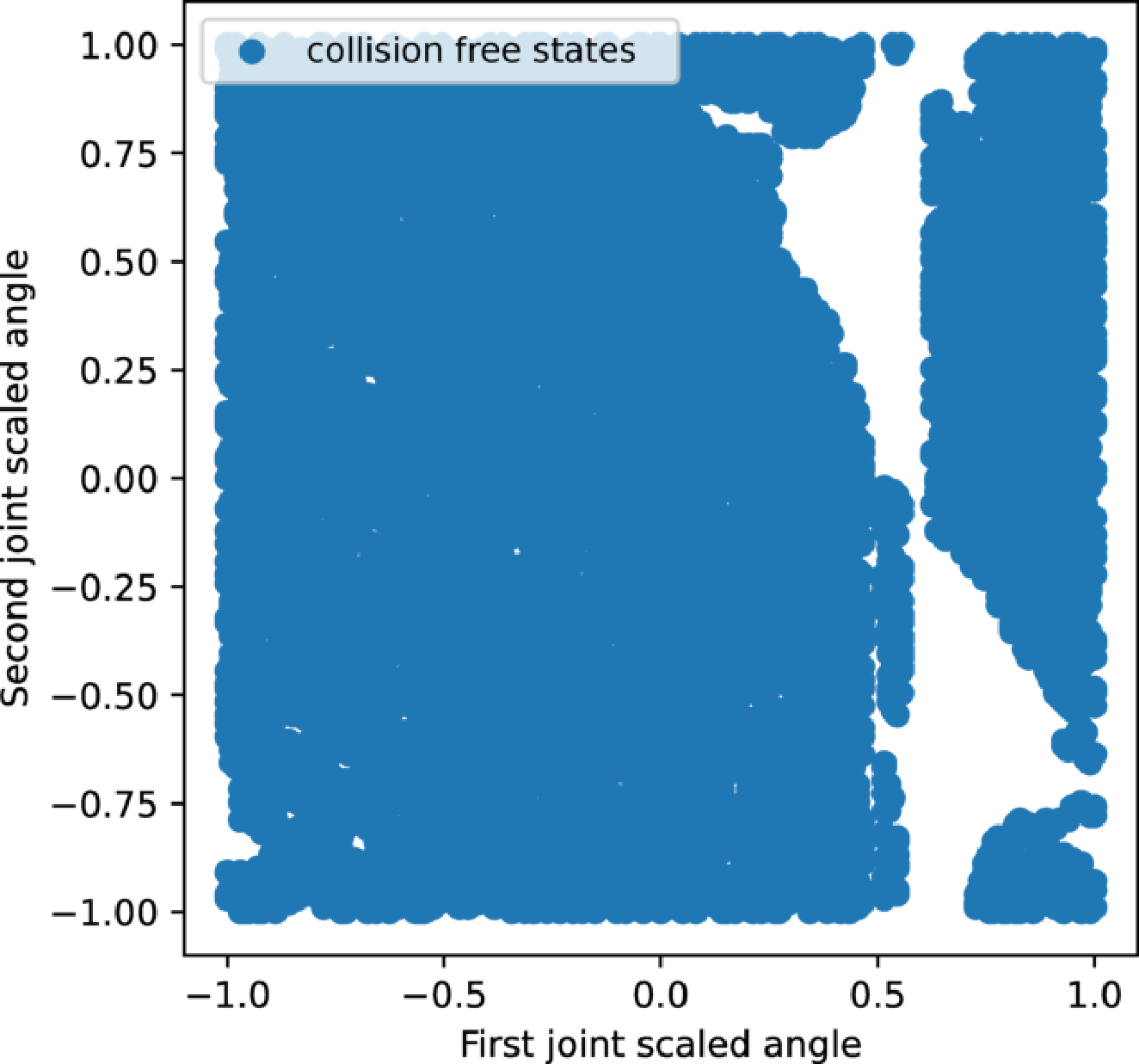

    \caption{CS where the obstacle constrains the first link of the manipulator}

    \label{fig:CS1comparisonGS}
\end{subfigure}
\caption{Two different CSs where the obstacle is positioned in two different places. Although they are graphically different, the GS indicates that they are topologically very similar (GS = 0.05975).}
\label{fig:comparisonGS}
\end{figure}

During our testing of GS on our dataset, we encountered a problem with the metric when obstacles partitioned the CS. According to the definition of GS in \eqref{eqn:GeomScore}, the score is close to 0 when two datasets have similar topological properties. However, in our test case involving a manipulator robot, when an obstacle blocks the first link, it generates a partition of the CS. The issue with this type of CS is that the GS will be very similar between CSs that do not have many holes but are partitioned. This is because the simplices are created separately or suddenly fuse into one component, as shown in \figref{fig:comparisonGS}.

\begin{figure}[ht]
\centering
\begin{subfigure}{0.45\columnwidth}
	\def\svgwidth{\textwidth}
    %
    %\import{#1.tex}
    %\def\svgwidth{\columnwidth}
    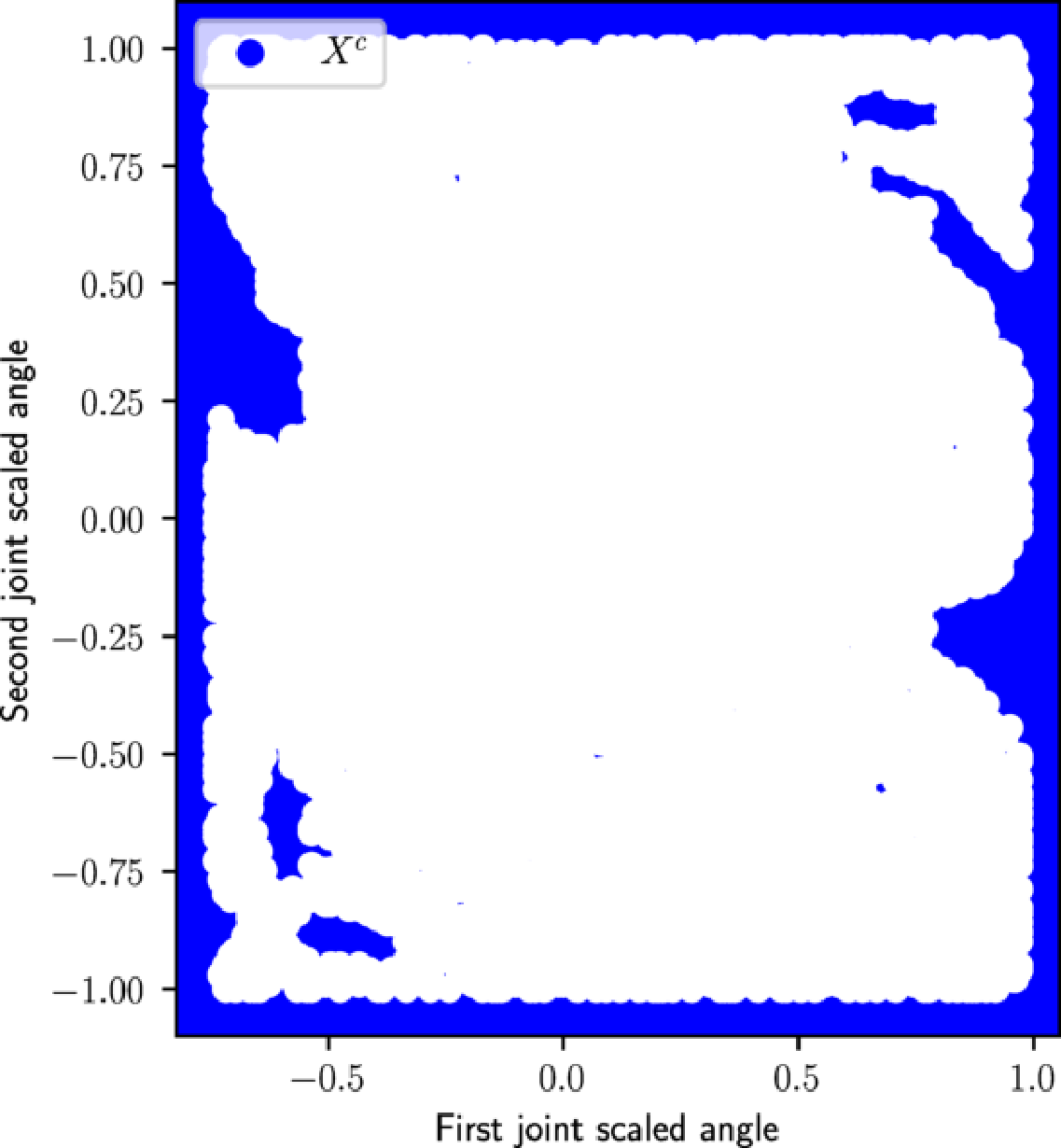

    \caption{$X_1$ represents a big hole in $X_1^c$.}

    \label{fig:CS0comparisonComplementGS}
\end{subfigure}
\hfill
\begin{subfigure}{0.48\columnwidth}
	\def\svgwidth{\textwidth}
    %
    %\import{#1.tex}
    %\def\svgwidth{\columnwidth}
    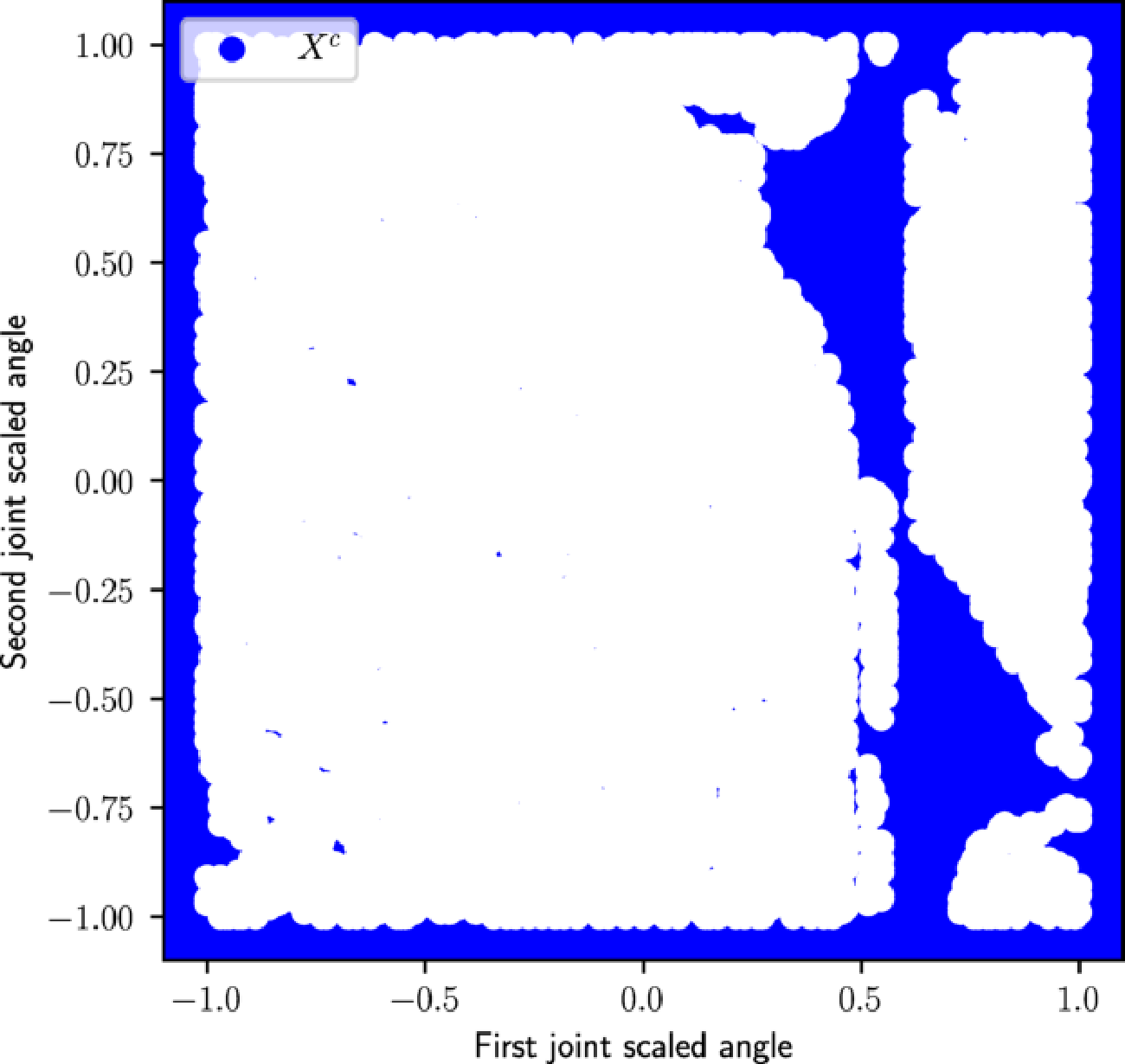

    \caption{$X_2$ represents several holes in $X_2^c$.}

    \label{fig:CS1comparisonComplementGS}
\end{subfigure}
\caption{By utilizing the complements of the point clouds, we can detect holes if CF-states are present.}
\label{fig:comparisonComplementGS}
\end{figure}

To address the issue of measuring similarity when there are no holes in the reconstructed CS, we propose incorporating the GS of the dataset's complement. By considering the complement, denoted as $\vector{X}^c$, we ensure that there is at least one hole if there is a CF state, which should be reflected in the GS estimation for the current $\vector{X}^c$. To illustrate, we show an example of obtaining the complement of the CS in \figref{fig:comparisonComplementGS}.

To obtain $\vector{X}^c$ we can retrieve the collision states that were produced during the sampling phase for obtaining $\vector{X}$. However, we also need to determine $\vector{X}^c$ for the generated points from the trained model. We suggest approximating $\vector{X}^c$ from $\vector{X}$ specifically in the context of manipulator robots, where only $\vector{X}$ is available.

 The algorithm that we propose operates under the assumption that the CS was sampled uniformly to partition the CS into smaller intervals for each dimension of the manipulator, denoted as $\vector{\theta}_i$, where $\vector{X} \subset \vector{\theta}^2$ and $i$ is the joint's vector entry number.

To increase the likelihood of capturing at least one point for each coordinate, the number of intervals is determined by the number of points originally sampled from the collision states. We divide the CS uniformly in function of the number of points of the dataset $\vector{X}$.

The next step is to determine whether there exists a predefined distance without any points in the current one-dimensional interval $\bm{\theta}_i$; from $\theta_{i,j,k}$ to $\theta_{i,j+1,k}$, where $j$ denotes the current partition of $\theta_{i}$ and $k$ the current partition of $\theta_{q}$, where $q\in \NaturalNumbers$. If there are no points in the  current interval, it could potentially represent a boundary of $\vector{X}$ from $\theta_{i,j,k}$ to $\theta_{i,j+1,k}$.

To verify that the empty interval is indeed where the states are in collision, we examine the neighboring intervals, specifically $\theta_{i,j,k+1}$ to $\theta_{i,j+1,k+1}$, to determine if they have a boundary that intersects with $\theta_{i,j,k}$. If an intersection exists, we conclude that we have found a boundary of $\vector{X}^c$. However, if there is no intersection, it is assumed to be a random distance between the cloud points generated from the sampling algorithm. For the two-dimensional case, we apply the algorithm to each entry $i$.

Once the boundary points of the joint are identified, we make a weighted random sampling of $\vector{X}^c$. The weight is defined by the area of the detected boundary regions. Algorithm \ref{alg:generationXc} outlines the procedure for generating $\vector{X}^c$ from $\vector{X}$.

\begin{algorithm}
    \caption{Generation of $\vector{X}^c$ from $\vector{X}$}
    \label{alg:generationXc}
    \KwData{$\vector{X} \subset \vector{\theta}^2$, $j \in \Real^+$ }
    \KwResult{
        $
        \vector{X}^c=[\mathbf{\vector{\theta}}^c_0,
        \mathbf{\vector{\theta}}^c_1]
        $
    }
    $intervals_j=\{[-\pi,-\pi+j),[-\pi+j,-\pi+2j),...,[\pi-j,\pi]\}$\;
    $itervals_{jk}=intervals_j\times intervals_j$\;
    $boundary_i=\emptyset$\;
    \For{$\vector{\theta}_i \in \vector{X}$}{
        \If{$i==1$}{
            $\vector{X}=\vector{X}^T$\;
        }
        \For{$interval_j \in intervals_{j_k}$}{
            \For{$interval \in intervals_j$}{
                \If{$isEmpty(\vector{X}(interval))$}{
                    $boundary_{i,j}.appends(\vector{X}(interval))$\;
                }
            }

        }

    }

        \For{$\vector{\theta}_i \in \vector{X}$}{
        \For{$boundary_j \in boundary_{i,j}$}{
            \If{$boundary_j\cap boundary_{j+1}$}{
                $boundaries.append(boundary_j,boundary_{j+1})$
            }
        }
    }
    $[\mathbf{\vector{\theta}}^c_0,\mathbf{\vector{\theta}}^c_1]=weightedRandom(boundaries)$\;
    \Return $[\mathbf{\vector{\theta}}^c_0,\mathbf{\vector{\theta}}^c_1]$\;

\end{algorithm}

\begin{figure}[ht]
\centering
\begin{subfigure}{0.45\columnwidth}
	\def\svgwidth{\textwidth}
    %
    %\import{#1.tex}
    %\def\svgwidth{\columnwidth}
    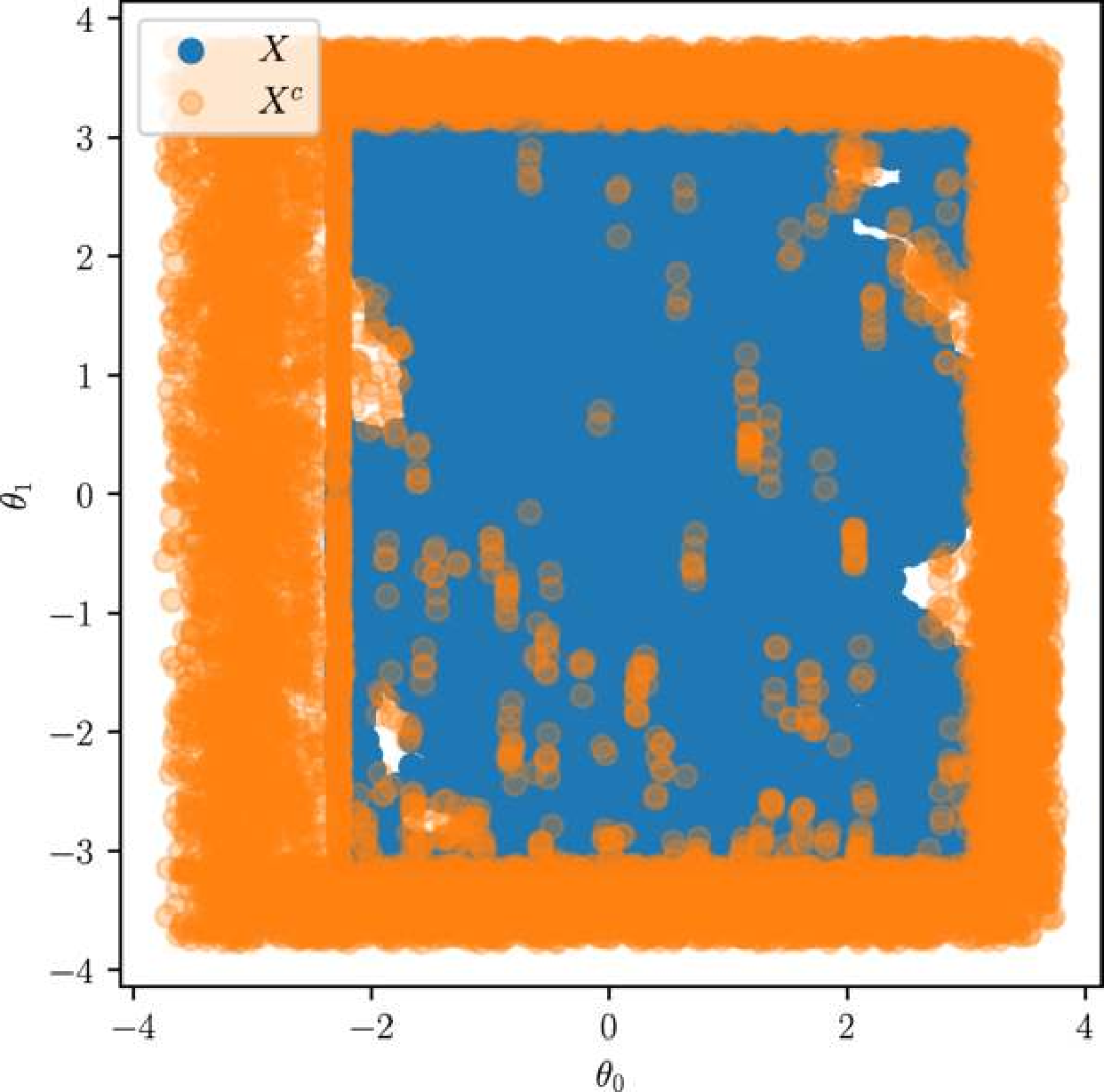

    \caption{Approximation of the complement of the CS $\vector{X}$}

    \label{fig:aproxComplement}
\end{subfigure}
\hfill
\begin{subfigure}{0.48\columnwidth}
	\def\svgwidth{\textwidth}
    %
    %\import{#1.tex}
    %\def\svgwidth{\columnwidth}
    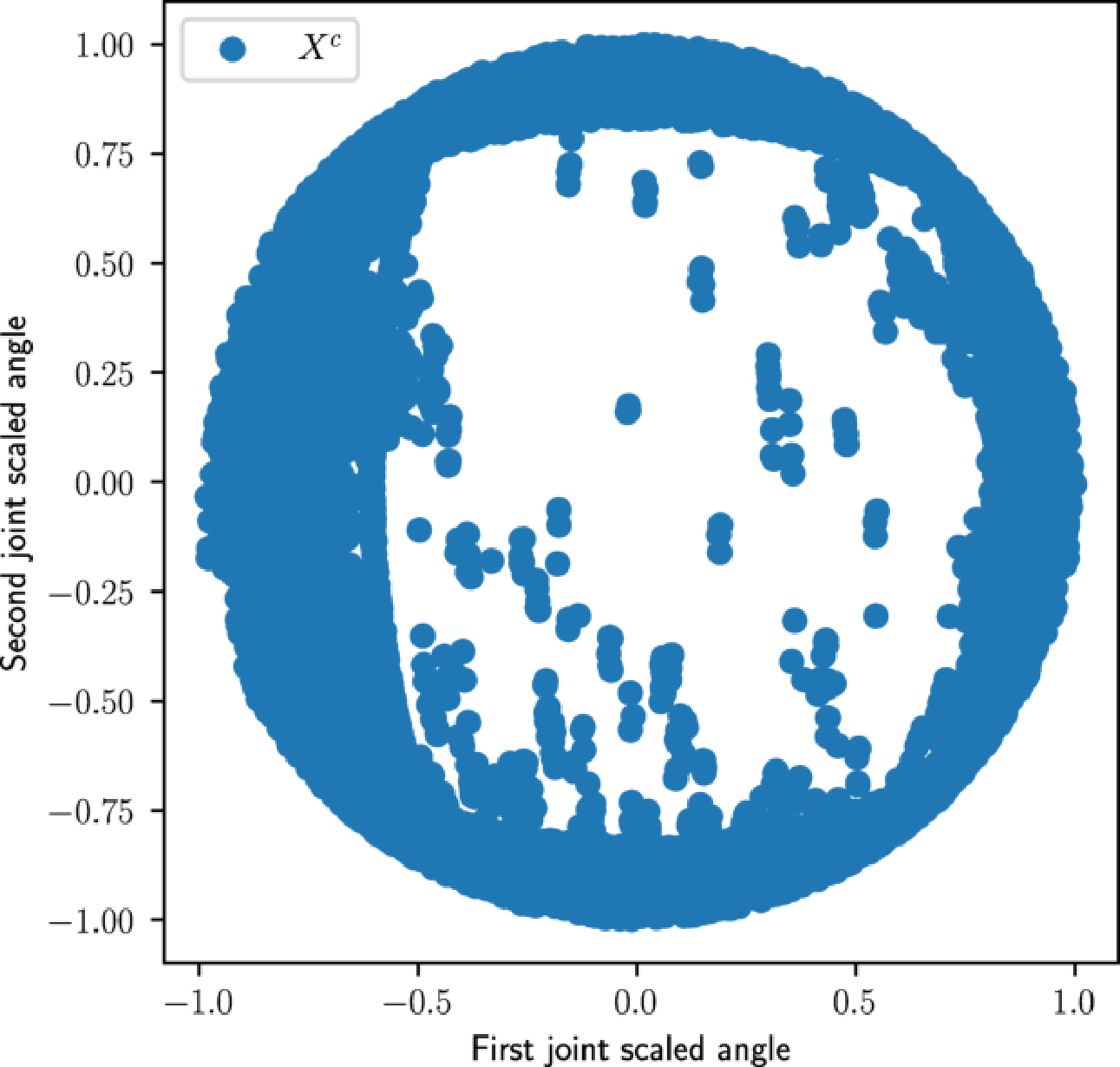

    \caption{Mapping from the square $\vector{X}^c$ to a circle}

    \label{fig:collisionStatesCircleTransform}
\end{subfigure}
\caption{After approximating the complement of CS (a) and mapping it (b), we obtain a new GS from the example of \figref{fig:comparisonGS} that reflects more the topological properties of the CSs, GS=0.244}
\label{fig:transformationComplement}
\end{figure}

In order to effectively apply the GS to our problem, we employed a mapping technique that involves converting the CS to a circular shape. This was a crucial step, as the GS relies on an open disk to define neighborhood, whereas our current representation has distinct, angular corners. By implementing this transformation, we were able to preserve the defining traits of the CS while ensuring that the same neighborhood definition could be applied. As a result, we were able to utilize the GS throughout the entire CF-square area, as demonstrated in \figref{fig:transformationComplement}.

To facilitate easier detection of the holes, we enlarged their size by applying the exponential function to the transformed circular data. This extension was applied to all data in the first quadrant, including the holes themselves. To ensure that the holes could be detected across all quadrants, we rotated the circle four times, each time with an angle of $\frac{\pi}{2}$. This rotation allowed us to define the new GS-C score given by:
\begin{align}
\resizebox{\columnwidth}{!}{$
\begin{aligned}
\text{GS-C}(\vector{X}_1,\vector{X}_2)=\frac{1}{2(t_{max}-1)} (\max_{GS}GS(\vector{R}(\Theta)\vector{X}_1,\vector{R}(\Theta)\vector{X}_2)\\+\max_{GS}GS(\vector{R}(\Theta)\vector{X}^c_1,\vector{R}(\Theta)\vector{X}^c_2))
    \label{eq:gs-c}
\end{aligned}
$}
\end{align}

with $\Theta\in \{0,\frac{\pi}{2},\pi,\pi+\frac{\pi}{2}\}$ and $\vector{R}$ is a rotation matrix.
\subsection{Clustering}

When it comes to path planning problems in robotics, the accurate sampling of CF states is of utmost importance. Any false positives can result in the robot colliding with its environment, causing potential damage and setbacks. By training a model to learn the topology of the CS, we can significantly enhance tasks related to path planning. This technique allows us to better understand the environment in which the robot is operating, allowing for more accurate and efficient planning of its path.

\begin{figure}
\begin{center}
\includegraphics[width=.5\textwidth]{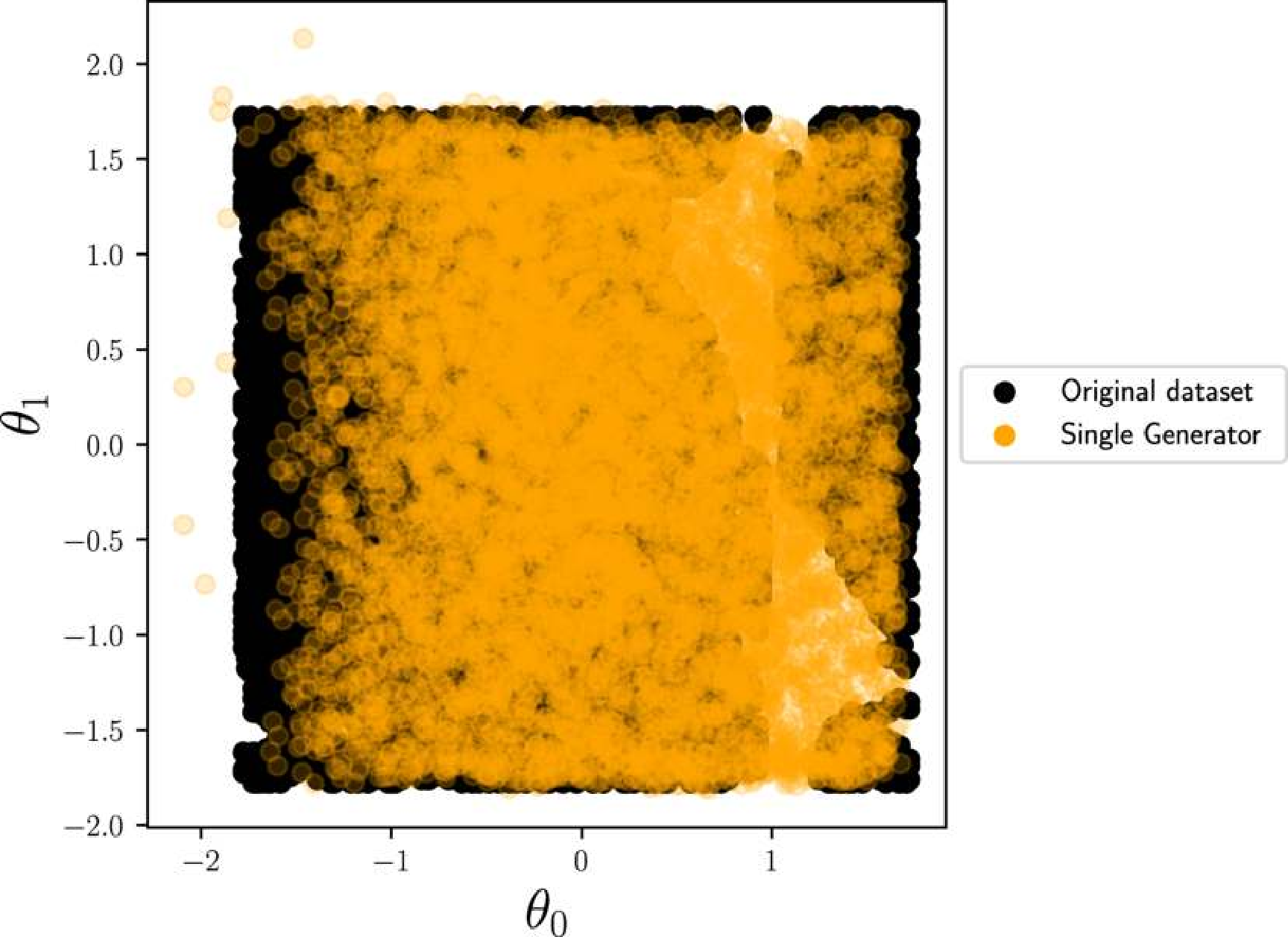}
\caption{After 500 epochs, the generator with only one cluster struggles to reconstruct the partition of CS. $\theta_i$ has been standardized.}
\label{fig:twoWorkinSpacesCSsOneGan}
\end{center}

\end{figure}

We encountered challenges when training a WGAN-GP with the entire dataset to reconstruct the $\StateSpace{X}_{free}$ of the robot. Specifically, generating holes or partitions of the CF states after changing the image-scenario conditioning proved difficult. This is not unexpected as the problem of learning a CS can be viewed as learning a distribution. When the conditioning splits the CS, a new distribution is created, which means the algorithm must effectively learn several distributions per condition. This scenario is illustrated in \figref{fig:twoWorkinSpacesCSsOneGan}.

To enhance the reconstruction of CSs, we suggest implementing a clustering approach. This approach draws inspiration from the field of image-to-image generation. The process involves training several local WGAN-GPs, each representing a different region of the image, alongside a set of corresponding critics. These critics and discriminators are then used to train a new generator that can generate more detailed images.

In the context of image-to-free-CS, clustering plays a critical role in reducing the overall complexity of the model. By decreasing the number of potential transformations of data points, clustering helps to streamline the process. For example, if a given cluster does not contain any holes in its conditioning, then the conditioning can be simplified to a straightforward translation of the point positions within the current configuration. This can greatly enhance the efficiency and effectiveness of the overall image-to-free-CS process.

To effectively partition the configuration space, we implemented k-means clustering with a value of k equal to 2 across the entire CS dataset. However, it is important to note that when the CS samples are diverse, the k-means algorithm will divide the data uniformly into 2 regions. The selection of 2 was based on testing, where we found that the algorithm was still able to effectively learn the conditioning of the CS with the size of the proposed dataset. While it may be possible to select more clusters, doing so could result in the curse of dimensionality if we have only a few samples for training each cluster using the WGAN-GP model. Therefore, selecting a larger number of clusters could increase the probability that the range of the homology group conditioned by the image-scenario remains constant, but it could also lead to issues with the curse of dimensionality.

During training, we encountered an issue where the non-clustered WGAN-GP was unable to effectively learn the conditioning with the encoder. We observed that some of the clusters would collapse when the encoder and WGAN-GP were trained simultaneously. Similarly, when training the model independently, it could not successfully learn the CS or conditioning. To address this issue, we introduced a reduction in complexity by training the generator to learn as if it were parameters of a Gaussian Mixture Model (GMM), which is defined by the following equation:
\begin{equation}
    p(x)=\sum^{K}_{k}p(x|z=k)p(z=k)
    \label{eq:gmm}
\end{equation}
where $p(x|z=k) \BehavesLike \Normal(\mu_k,\Sigma_k)$.

Previous works such as \cite{10.1007/978-3-319-47437-3_4} and \cite{9712347} have GMMs for path planning with RRT-based algorithms to learn the biased configuration space directly. However, the challenge with this approach is the difficulty in conditioning the GMM to changes in the initial and final states of the path.

In contrast, our method represents the initial and final states as a concatenation of the latent vector for generation, which simplifies the problem. We use Gaussian models to approximate the configuration space, with only the means of a normal distribution ($\mu_k$) being learned by the generator, and $\Sigma_k$ fixed as a constant diagonal matrix with $\sigma=0.025$. We selected $\sigma$ empirically, testing the largest possible value that could significantly reduce the search space while still providing an accurate approximation of the configuration space. 

Our approach can be viewed as injecting Gaussian noise into the data to prevent the model from collapsing to specific regions instead of approximating the real probability distribution, as suggested in \cite{6795935}. By adopting this approach, we were able to improve the model's ability to learn the conditioning and enhance the overall effectiveness of the training process. To be able to estimate the gradient for the normal distributions, we use the reparametrization trick from \cite{Kingma2014}.

We conducted experiments to generate multiple sets of parameters for the Gaussian distributions per query, and we increased the output dimension of the generator if that was necessary. Our results indicated that incorporating the mean of the discriminator with all the means generated by each iteration improved the fitting of the training data. However, we observed that some of the means fell outside of the configuration space. We concluded that the critic loss was responsible for generating outliers that extended beyond the boundaries of the clusters.

In order to translate the predicted means that are outside of the training dataset for reconstructing the configuration space, we employed an iterative approach using a smooth approximation of the maximum function \cite{9878772}. We performed this process for multiple values, taking the smooth maximum of the critic and directly adding it to the loss function during each iteration. This allowed us to achieve a more accurate and robust reconstruction of the configuration space, even for the predicted means that fall outside of the training dataset.
\begin{equation}
    \mathcal{S}_{\epsilon}(x_1,x_2)=\frac{(x_1+x_2)\sqrt {(x_1-x_2)^2+\epsilon^2}}{2}
\end{equation}
$\mathcal{S}_{\epsilon}\rightarrow \max$ when $\epsilon \rightarrow 0$.

Our proposed WGAN-GP algorithm, with multiple clusters; MultiWGAN-GP can be expressed as Algorithm \ref{alg:mgan}.

\begin{algorithm}
  \caption{Generating means from clusters}
    \label{alg:mgan}
    \KwData{$\vector{X} \subset \vector{\theta}^2,\vector{y}\subset I:U\rightarrow [-1,1]^{ 3\times 48 \times 48}$ as subset of the input images, $l$ number of clusters, $\vector{x}^{(j)} \subset \vector{X}$ is a subset of the training data of the $j$th cluster, $K$ is number of means per cluster, $m$ is batch size, $n_{critic}$ is number of iteration of the critic, and $q(\vector{z}|\vector{y})$ is previously trained encoder}
    \KwResult{$g_{\rho}^{(0...l)},f_{w}^{(0..l)}$}
  \For{$j= 0:l$}{
    \While{$\rho^{(j)}$ has not converged}{
        \For{$t=0,...,n_{critic}$}{
            Sample $\{\vector{x}^{(i,j)}\}^m_{i=1}\BehavesLike \Probability_r$ a batch of size $m$ from the cluster $j$ from the real data\;
            Sample $\{\vector{z}^{(i,j)}\}^m_{i=1}\BehavesLike q(\vector{z}|\vector{y}^{(i,j)})$ a batch from a previous trained encoder\;
            $\tilde{\vector{x}}^{(i,j)}_{0..K}\leftarrow \Normal_{0..K}(g_\rho^{(j)}(\vector{z}^{(i,j)}))$\;
            $L^{(i,j)}\leftarrow  [\frac{1}{m} \sum^m_{i=1}f^{(j)}_w(\vector{x}^{(i,j)})-\frac{1}{m}\sum^m_{i=1}f_w^{(j)}(\tilde{\vector{x}}^{(i,j)}_{0..K})+\frac{\lambda}{m}\sum^m_{i=1}(\Norm{\Gradient_{\hat{\vector{x}}^{(i,j)}}f_w^{(j)}(\hat{\vector{x}}^{(i,j)})}-1)^2-\mathcal{S}_\epsilon(f_w^{(j)}(\tilde{\vector{x}}^{(i,j)}_{0..K}))]$\;
            $w^{(j)}\leftarrow Adam(\Gradient_{w^{(j)}} L^{(i,j)},w^{(j)})$

        }
        Sample $\{\vector{z}^{(i,j)}\}^m_{i=1}\BehavesLike q(\vector{z}|\vector{y}^{(i,j)})$ a batch from the encoder\;
        $\tilde{\vector{x}}^{(i,j)}_{0..K}\leftarrow \Normal_{0..K}(g_\rho^{(j)}(\vector{z}^{(i,j)}))$\;
        \resizebox{0.8\columnwidth}{!}
     {$\rho^{(j)} \leftarrow Adam(\Gradient_{\rho^(j)}[-\frac{1}{m}\sum_{i=1}^{m}f_w^{(j)}(\tilde{\vector{x}}^{(i,j)}_{0..K})],\rho^{(j)})$\;}

    }
  }
\end{algorithm}

For the final step, we utilize the trained critics from Algorithm \ref{alg:mgan} to bias the models with the local critics of each cluster $f^{(j)}_w$ using the whole dataset. This approach can improve the performance of the general WGAN-GP model, and help it converge closer to the actual training data in the local regions. To implement this last step to train the global GAN, we modify lines 7 and 11 of Algorithm \ref{alg:mgan} by incorporating the discriminator of the clusters. Specifically, we add the local trained critic of each cluster to the loss function of the global GAN as follows:
\begin{multline}
    L^{(i)}\leftarrow \Gradient_{w}[\frac{1}{m} \sum^m_{i=1}f_{w}(\vector{x}^{(i)})-\frac{1}{m}\sum^m_{i=1}f_{w}(\tilde{\vector{x}}^{(i)})\\+\frac{\lambda}{m}\sum^m_{i=1}(\Norm{\Gradient_{\hat{x}^{(i)}}D(\hat{x}^{(i)})}-1)^2]\\
      \rho \leftarrow Adam(\Gradient_{\rho}[-\frac{1}{m}\sum_{i=1}^{m}f_{w}(\tilde{\vector{x}}^{(i)})-\frac{1}{l} \sum^l_{j=1}\frac{1}{m}\sum_{i=1}^{m}f^{(j)}_w(\tilde{\vector{x}}^{(i)})],\rho)
    \label{eq:multiDirector}
\end{multline}
with $l=1$ and $\sigma=0$ for the global critic and generator training.

This modification ensures that the general model is also guided by the local characteristics of each cluster, resulting in better performance and more accurate results.

\subsection{Planner}

To guide the path towards the desired region in the free configuration space, we utilized a technique inspired by \cite{Wang2020NeuralRL}. Our approach involved using the learned generator as a sampler for RRT path planning algorithm. However, we made a modification in our implementation to specifically handle situations where our encoder failed to capture the complete encoding of a previously observed scenario. To this end, we increased the value of $\vector{\sigma}$ directly from the encoder's output, instead of trying to find a ratio between sampling from the uniform distribution and the generator. This idea is similar to having a forward trajectory of $k \in \NaturalNumbers$ diffusion steps from the predicted distribution $f_w(\cdot)$ with fixed mean $\vector{\mu}\neq \vector{0}$ \cite{10.5555/3045118.3045358}.

\begin{algorithm}
    \caption{RRT with MultiWGAN-GP}
    \label{alg:biasedRRT}
    \KwData{$x_{init},x_{goal},\vector{y}\subset I:U\rightarrow [-1,1]^{t\times 3\times 48 \times 48},t \in \NaturalNumbers$ is the number of points to sample from MultiWGAN-GP with $t\geq n$ iterations, $\vector{\epsilon_\sigma}=[0,\epsilon_1,...,\epsilon_k], k\in \NaturalNumbers, \epsilon_i \in \Real^+$ and $\epsilon_i>\epsilon_{i-1}$ is the amount of the $k$ perturbation of $\vector{\sigma}$. }
    \KwResult{
        $
        G
        $
    }
    $V\leftarrow x_{init},E\leftarrow \emptyset$\;
    $\vector{\mu,\sigma} \leftarrow q(\vector{z}|\vector{y})$\;
    $BiasSampler=\cup^{k}_{i=0} f_w(\Normal(\vector{\mu,\sigma}+\vector{\epsilon_\sigma[i]}))$\;
    \For{$i=1,...,n$}{
        $x_{rand}\leftarrow BiasSampler[i]$\;
        $x_{nearest}\leftarrow Nearest(G=(V,E),x_{rand})$\;
        $x_{new}\leftarrow Steer(x_{nearest},x_{rand})$\;
        \If{$ObstacleFree(x_{nearest},x_{new})$}{
            $V\leftarrow V \cup \{x_{new}\}$\;
            $E\leftarrow E \cup \{(x_{nearest},x_{new})\}$\;
        }

    }
    \Return $G=(V,E)$\;
\end{algorithm}

Our objective in taking this approach was to cover a more diverse set of latent vectors, thereby helping the generator to incorporate images that exhibit fewer similar features. This method can prove useful in situations where the encoder's trained query is unable to find any CF states within a predetermined number of generator samples. By employing this method, it becomes possible to identify CF points, even in scenarios where no examples closely resemble the trained data by leveraging shared characteristics among multiple training data scenarios; this implementation is reflected in Algorithm \ref{alg:biasedRRT}.

\section{Experimental results}
In order to provide visual representation of the results, we have designed a setup that incorporates a two-degree-of-freedom manipulator robot. The purpose of this setup is to demonstrate, through graphical means, the effectiveness of our proposed architecture in learning a two-dimensional configuration space that is dependent on the position of an obstacle. Furthermore, we aim to demonstrate how our architecture can be applied to diverse problems that necessitate understanding of the configuration space.

All the models are trained on a system with 2 x Intel Gold 6148 Skylake, 16 GB of RAM and 2 x NVidia V100SXM2. For deployment, we use Ubuntu 22.04 running on a 3.60 GHz × 8 Intel Core i7-9700K processor, 16GB RAM on NVidia RTX 2070.

We propose two sets of experiments to examine the effectiveness of partitioning the original dataset into clusters and learning the parameters of Gaussian distributions versus training the WGAN-GP with the image encoder directly. The training dataset comprises 100 image-scenarios and their corresponding configuration spaces of a simulated 2-dimensional 2-DOF manipulator robot with circular obstacles randomly placed within its working space. All input images of the scenario were resized to 48 x 48 pixels. We utilized the Pytorch Lightning framework with Adam optimizer parameters derived from \cite{10.5555/3295222.3295327} for training. Our experiments hyperparameters consisted of a learning rate of $4\mathrm{e}{-5}$, a batch size of 512, a regularization coefficient $\lambda$ of 10, an encoder regularization coefficient, $K=$4 means per cluster, a latent dimensionality of 512 for $z$, and $n_{critic}=5$ training iterations per generator iteration. All the CSs are standarized. Our code is openly available\footnote{\url{https://bitbucket.org/joro3001/multiwgangp/}}.

We employ an image to represent the conditioning factor in our experiment. This image encapsulates the obstacle's representation within the robot's operational space. To streamline the experimentation process, we opted for three circles of radius one unit to serve as the obstacle. Additionally, we include the robot's starting position, which is at the initial state of $\vector{0}$ before the path is planned.

Please refer to Appendix \ref{FirstAppendix} for a comprehensive description and visual representation of the encoder, decoder, generator, and critic components.

\subsection{Reconstruction of the free-CS}
The aim of this experiment is to demonstrate the effectiveness of our proposed architecture in reconstructing a CS in comparison to using the entire dataset. To achieve this, we utilize a limited dataset and evaluate both a simple WGAN-GP model and our advanced MultiWGAN-GP model. Through this comparison, we showcase the superior performance of our proposed architecture.

We conducted training for both models over 400 epochs. In the case of the MultiWGAN-GP model, we opted to group the data into two distinct clusters to minimize the likelihood of gaps within each cluster. We experimented with using additional clusters initially, but found that reducing the number of samples per cluster resulted in the curse of dimensionality, which adversely affected the overall training process.

\begin{figure}
\begin{center}
\includegraphics[width=.5\textwidth]{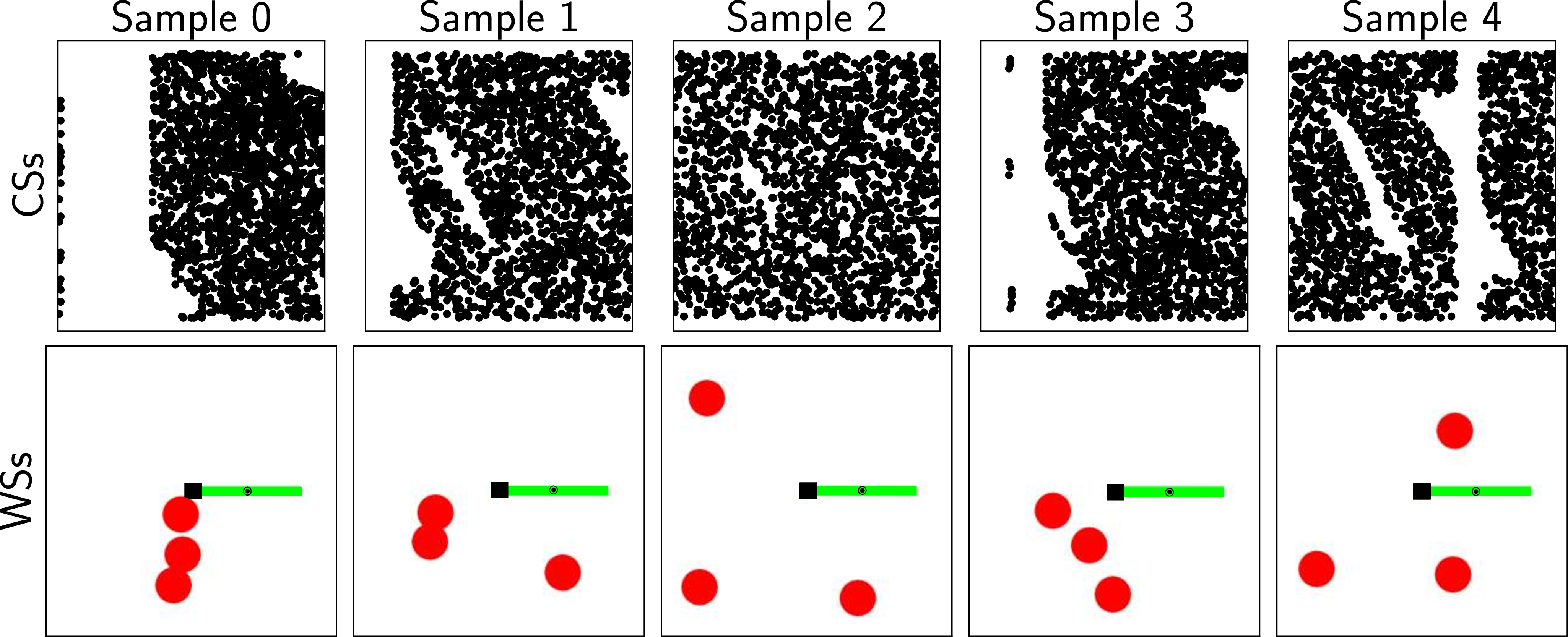}
\caption{Five samples from the dataset for training.}
\label{fig:dataset}
\end{center}

\end{figure}

We generated 100 random CSs where the image-scenario have 3 circular obstacles positioned randomly with a 2-dimensional 2-DoF manipulator  as training data. Each CS consist of 10000 random samples of CF states. We can see some samples of the dataset in \figref{fig:dataset}.

\begin{figure}
\begin{center}
\includegraphics[width=.5\textwidth]{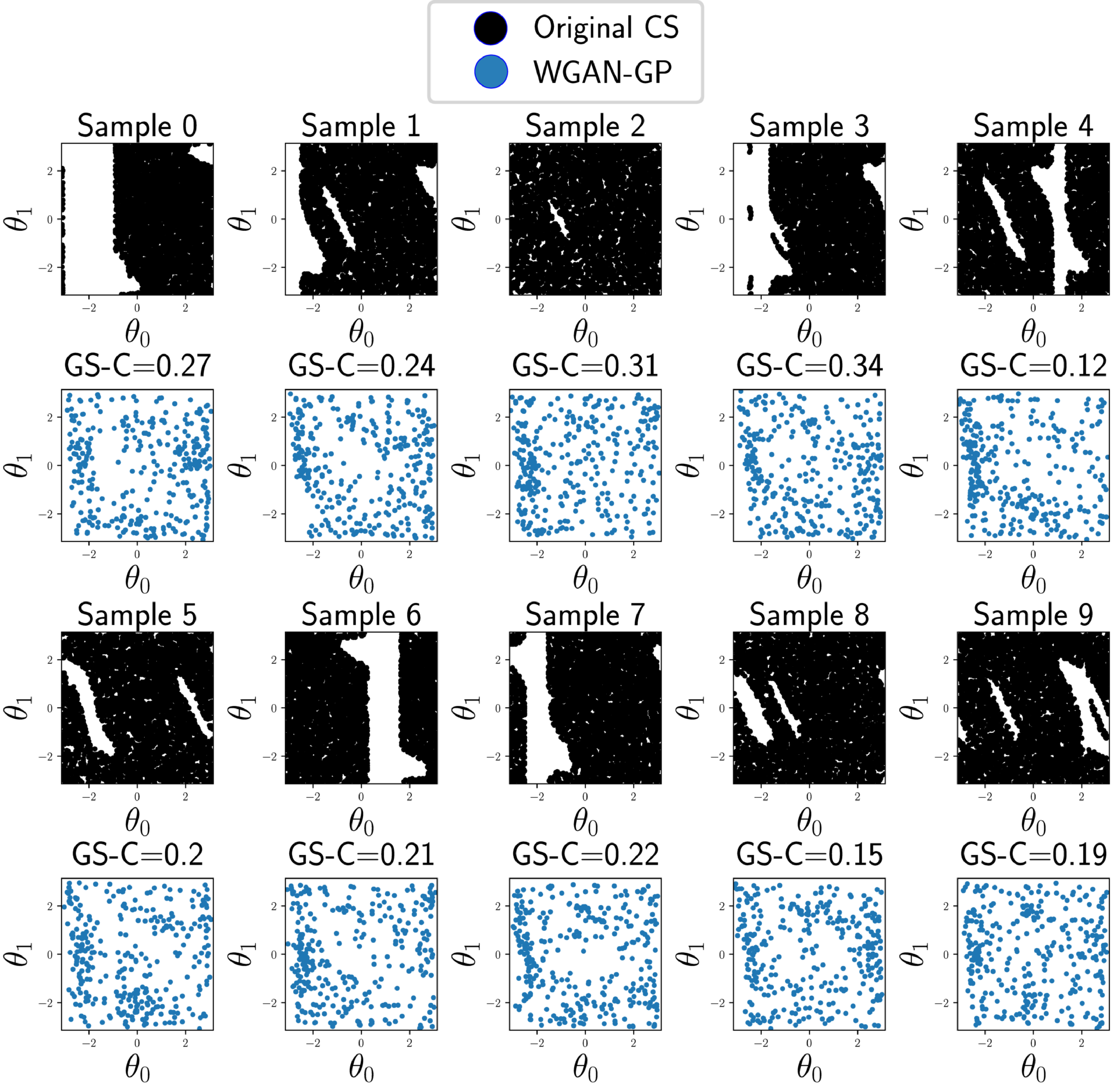}
\caption{Reconstruction of the CS using WGAN-GP directly from Eq. \eqref{eq:generatorAndVAE}. Lower GS-C is better.}
\label{fig:simpleModel}
\end{center}

\end{figure}

\figref{fig:simpleModel} illustrates the results of reconstructing the training data using Eq. \eqref{eq:generatorAndVAE}. The GS-C scores and plot reveals that the models are dissimilar, as the generated states fail to capture the variety present in the data, and instead appear to be centered around the mean of all the models. While the model is able to learn the data boundaries standarized between $[-\pi,\pi]$, it fails to differentiate between CF states and those that are in collision, resulting in collisions being generated. This failure can be attributed to the encoder training being incorporated within the WGAN-GP training process. Specifically, the cost defined by Eq. \eqref{eq:kl} converges close to the normal distribution, thereby impacting the WGAN-GP's ability to learn any conditioning. Even with offline VAE training, the WGAN-GP still struggles to learn the configuration states.

\begin{figure}
\begin{center}
\includegraphics[width=.5\textwidth]{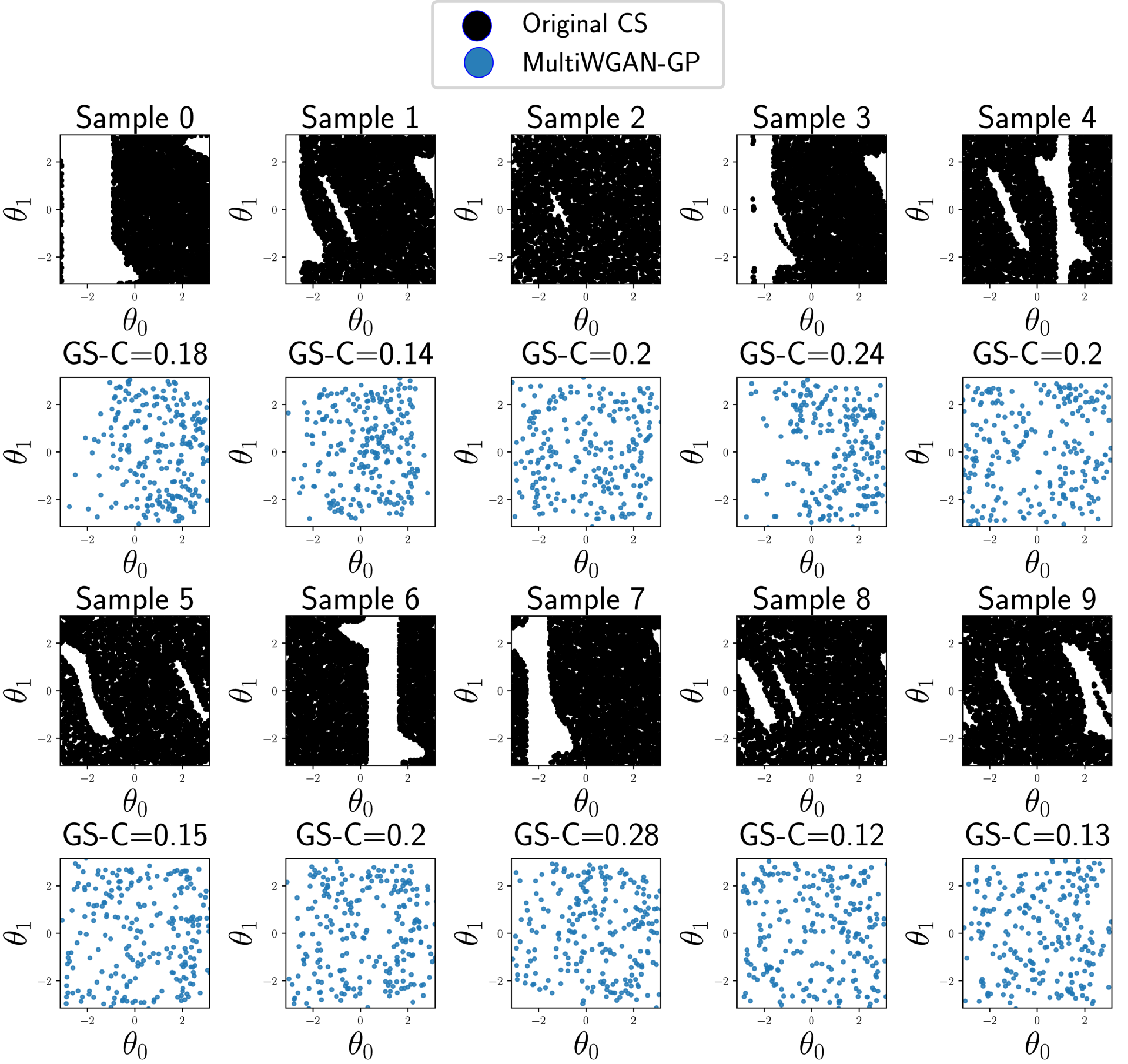}
\caption{Reconstruction of the CS with MultiWGAN-GP one cluster.}
\label{fig:resultsReconstructionClusters}
\end{center}
\end{figure}

\figref{fig:resultsReconstructionClusters} depicts the reconstruction of the training data using our proposed MultiWGAN-GP technique. The results demonstrate the efficacy of our method, as only one cluster is required to generate the data. The GS-C analysis indicates that the worst-case scenarios are avoided, and we can visually observe that the conditioning from the scenario with the encoder is being learned, despite the fact that the encoder was trained independently from the WGAN-GP in this case. This approach offers several advantages, such as accelerating the training of the WGAN-GP by independently training the encoder, and utilizing the same encoder to train each cluster when using the multi-cluster approach with MultiWGAN-GP.

\begin{figure}
\begin{center}
\includegraphics[width=.5\textwidth]{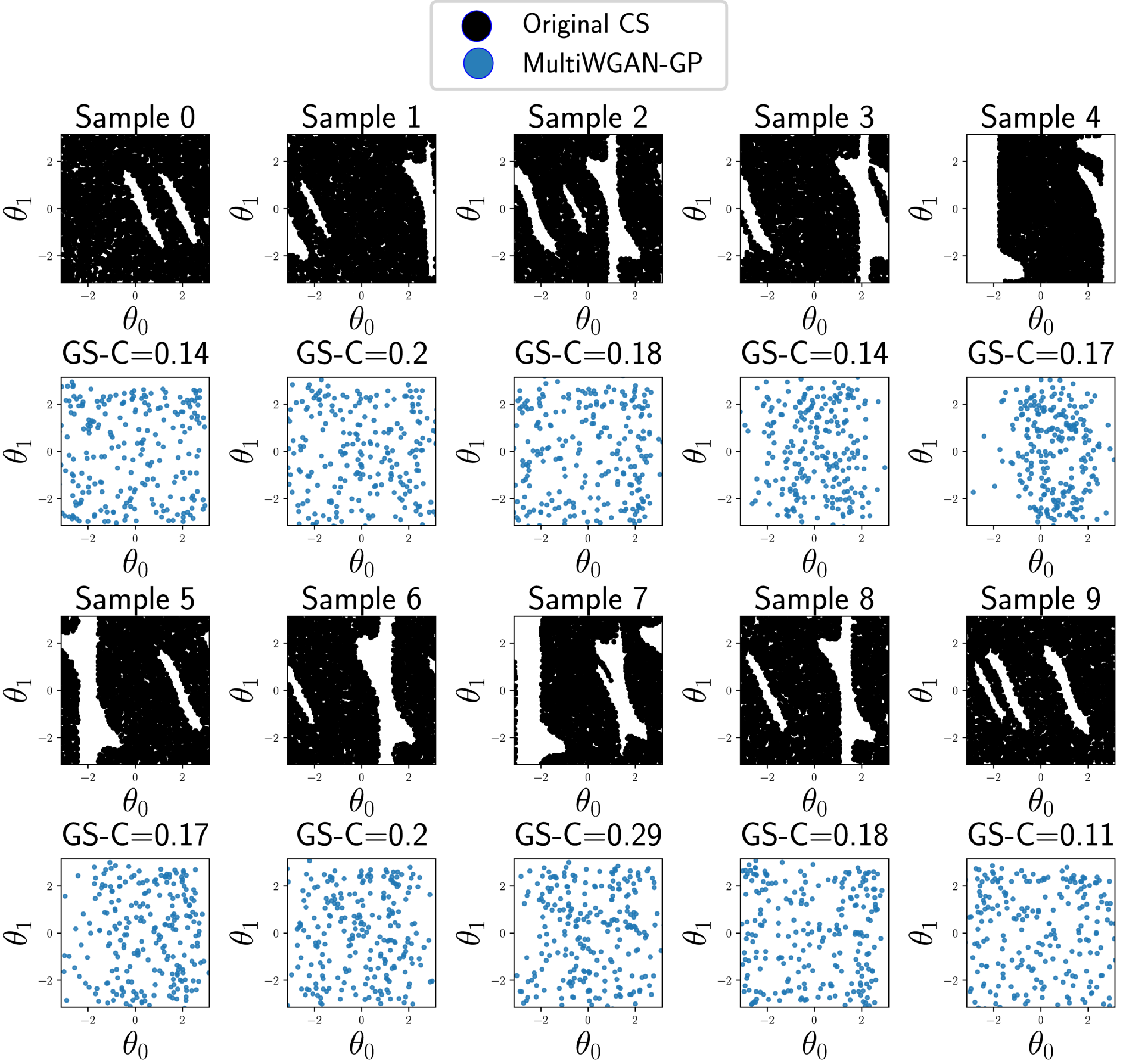}
\caption{Reconstruction of the CS with MultiWGAN-GP one cluster with non previously seen data.}
\label{fig:resultsExtrapolationOneCluster}
\end{center}
\end{figure}

\figref{fig:resultsExtrapolationOneCluster} illustrates that when testing the model with configuration spaces that have not been previously observed, the resulting approximation is very similar to the training data from \figref{fig:resultsReconstructionClusters} when the CSs are similar to those used to train the encoder. However, in cases where the configuration spaces are significantly different from those used during training or have small holes inside the CS, the generator is unable to accurately predict the CS. We conclude with this finding that we need to incorporate a more diverse range of obstacle positions during training to account for a wider variety of CS scenarios, specially when the collision area is relatively small.

\begin{figure}
\begin{center}
\includegraphics[width=.5\textwidth]{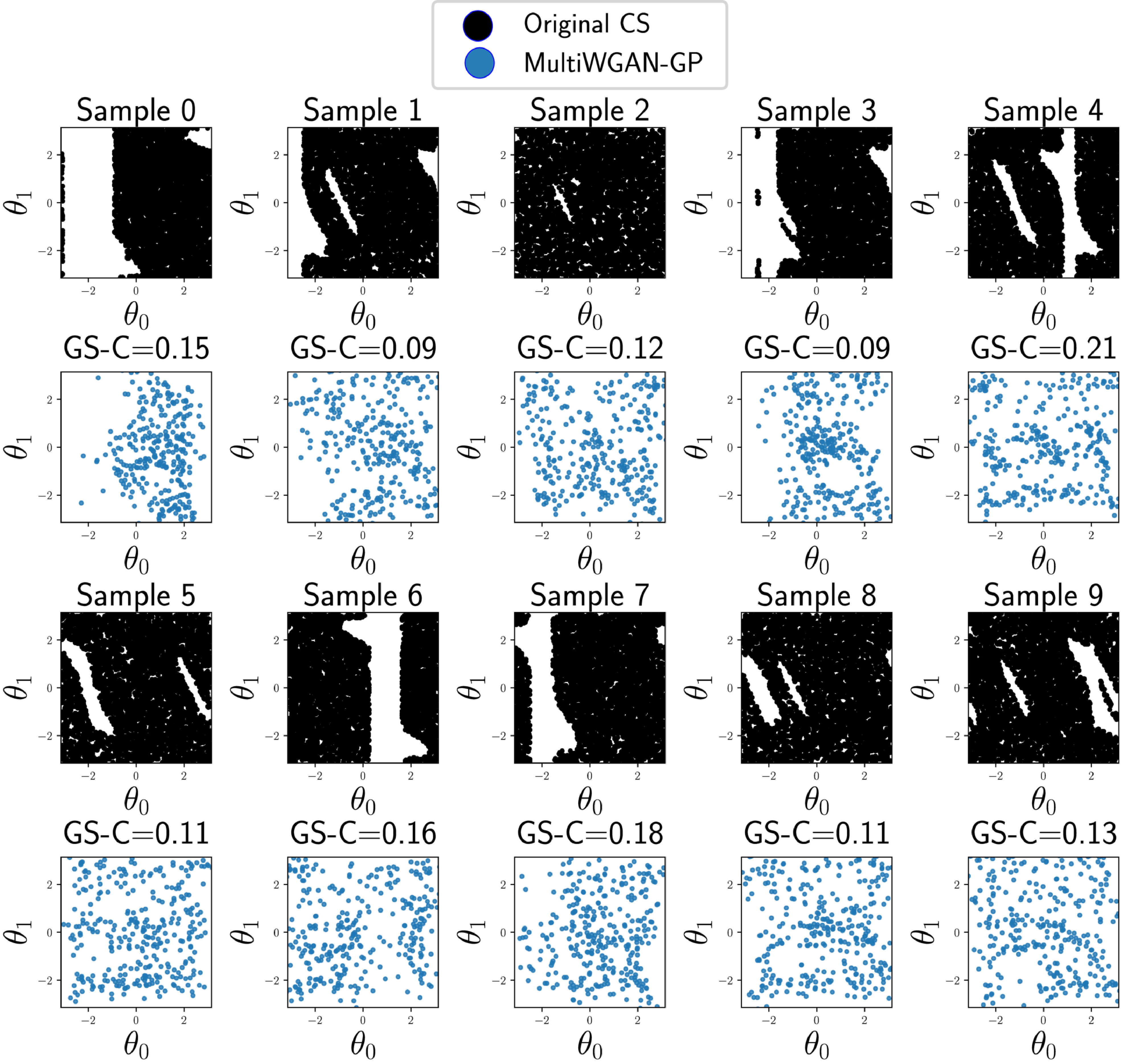}
\caption{Reconstruction of the CS with MultiWGAN-GP 2 clusters. Visually similar to one cluster reconstruction but the GS-C is improved.}
\label{fig:resultReconstructionMultiOnly}
\end{center}

\end{figure}

Training with MultiWGAN-GP and multiple clusters, we can see from the results shown in \figref{fig:resultReconstructionMultiOnly} that the improvement in GS-C is significant compared against one cluster.

\begin{figure}
\begin{center}
\includegraphics[width=.5\textwidth]{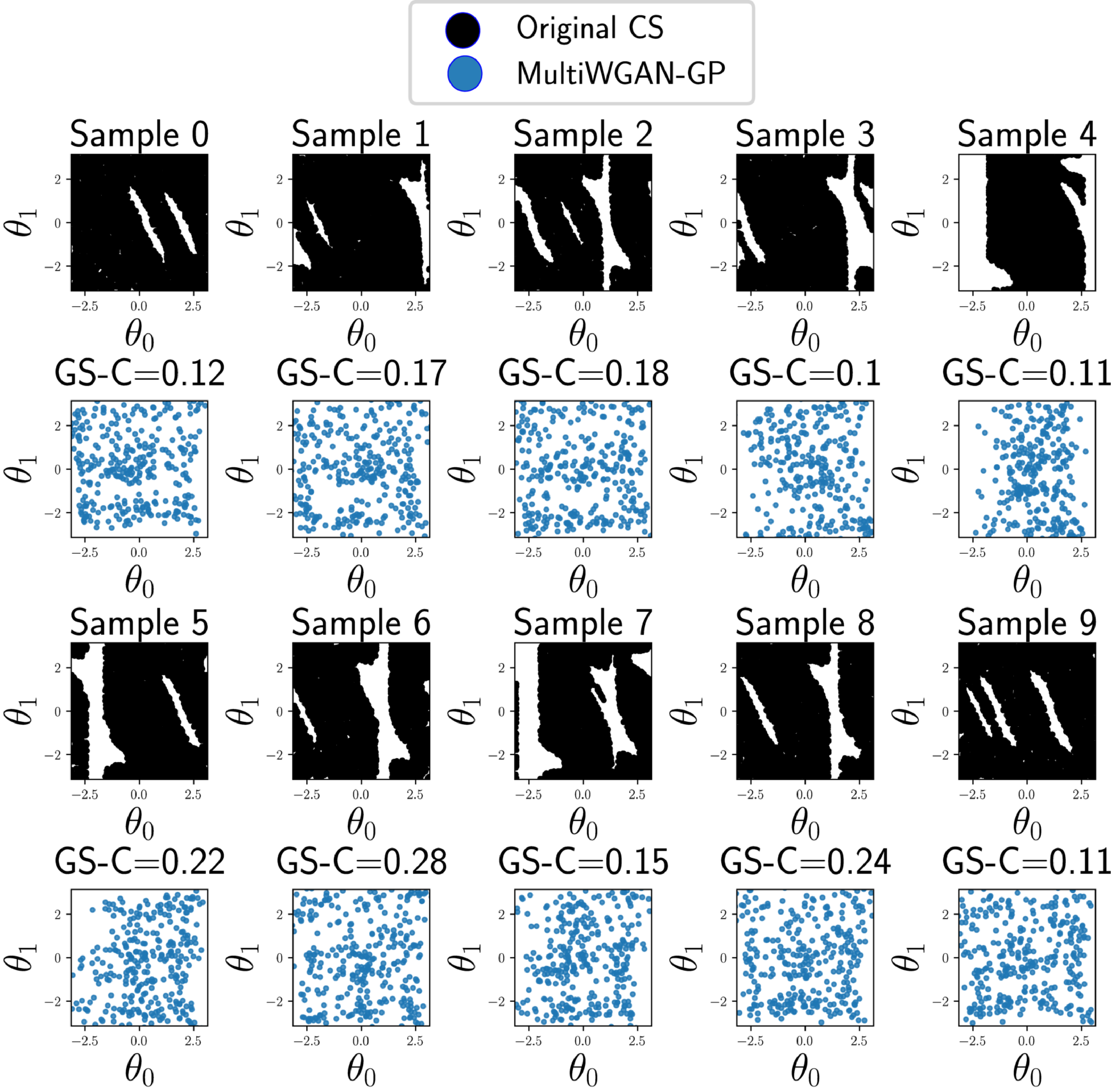}
\caption{Reconstruction of the CS with MultiWGAN-GP 2 clusters with previously unseen data.}
\label{fig:multipleClusterExtrapolation}
\end{center}

\end{figure}

Regarding extrapolation, we can observe from \figref{fig:multipleClusterExtrapolation} an improvement in extrapolation to previously unseen scenarios in most cases, compared to the MultiWGAN-GP with only one cluster.

\subsection{Path planning with extra models}

To train the generator for sampling paths in the $\mathbb{S}^1 \times \mathbb{S}^1$ configuration space, we first transformed the data by embedding it into a 4-dimensional Euclidean space, where we estimated the sine and cosine of each angle $\theta_i$. This allowed us to represent each path as a continuous line and eliminate any discontinuities in the data.

To address the variability of the paths in the RRT algorithm, we used RRT* paths as training data, which provided a more stable distribution that did not fluctuate significantly when the configuration space was changed. In our experiments, we allowed the RRT* algorithm 10 seconds to find the path and rewire the tree. We used path length as the minimization objective for the generator training, as it provides a reliable measure of the quality of generated paths.

In the previous section, we generated 10 new different scenario/configuration spaces for testing, and in this section, we estimated paths in each of these spaces. Our objective was to combine the same critics used to learn the CS reconstruction and improve the RRT planner's quality. To generate more data and overcome the curse of dimensionality, we interpolated all the paths.

In our proposed model, we used the same auto-encoder and parameters that were used in the CS learning phase. To evaluate our approach's improvement compared to other methods, we trained two WGAN-GP paths: one using only the auto-encoder and another using the two critics from the previous section. Both models were trained for 600 epochs.

\begin{figure}[tb]
\centering
\begin{subfigure}{0.49\columnwidth}
	\def\svgwidth{\textwidth}
    %
    %\import{#1.tex}
    %\def\svgwidth{\columnwidth}
    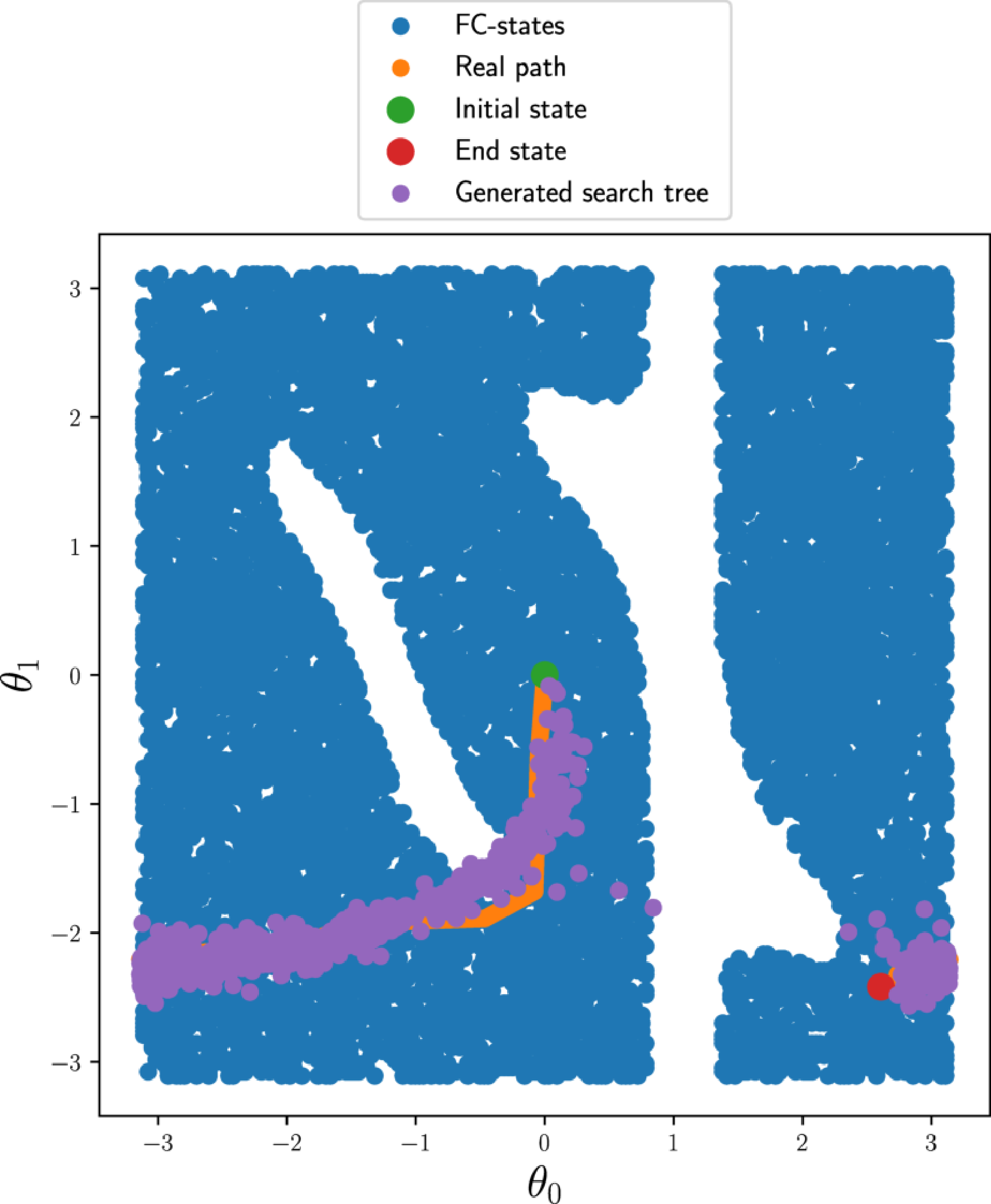

    \caption{Sampling using the model trained  with only the critic for the path.}

    \label{fig:samplingPath}
\end{subfigure}
\hfill
\begin{subfigure}{0.49\columnwidth}
	\def\svgwidth{\textwidth}
    %
    %\import{#1.tex}
    %\def\svgwidth{\columnwidth}
    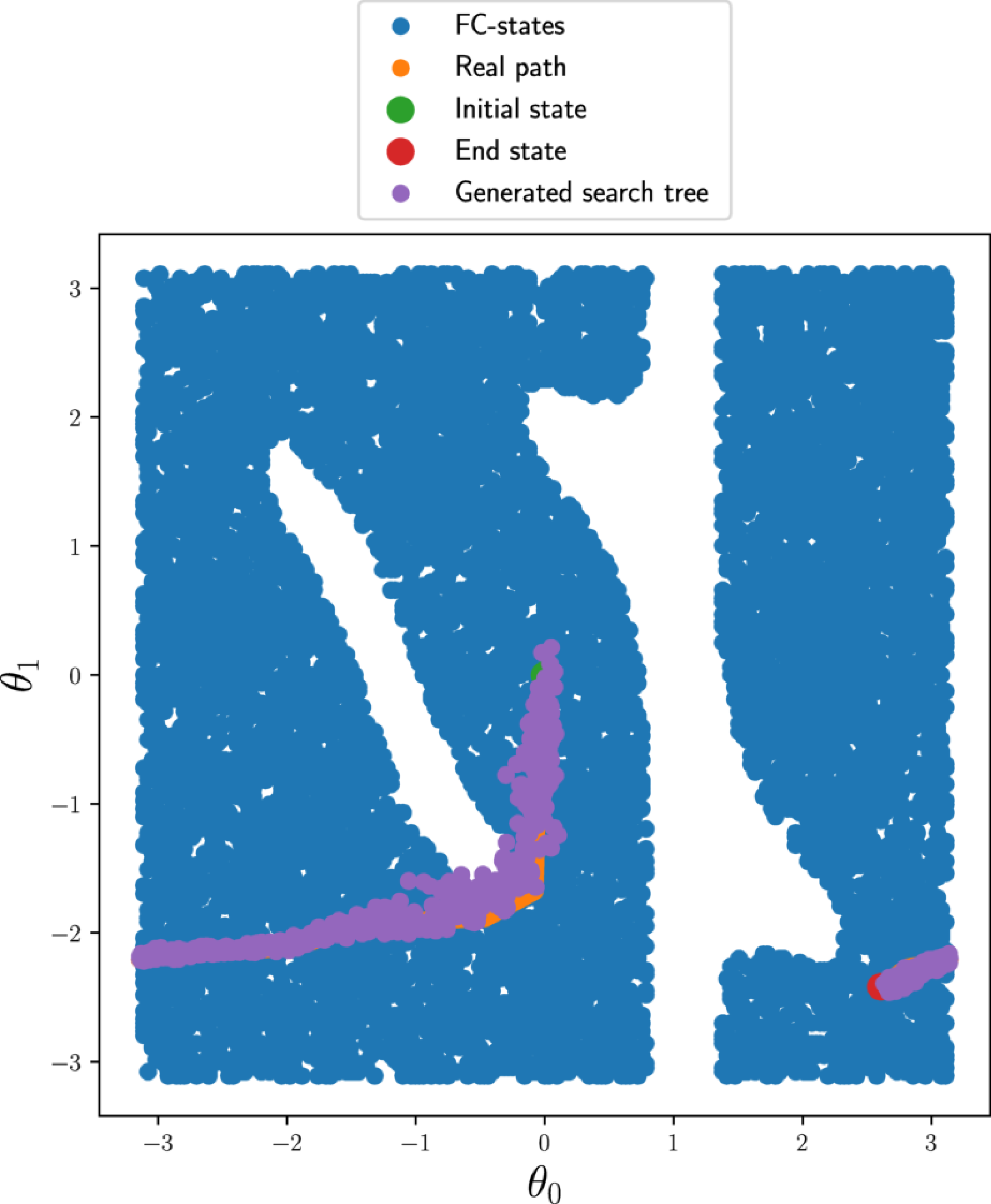

    \caption{Sampling using the model trained  with the critics used to learn the CS and path.}

    \label{fig:samplingPathMulti}
\end{subfigure}
\caption{Testing the sampling from the generators used for sampling. }
\label{fig:comparisonPathsSampling}
\end{figure}

As expected, our proposed approach of incorporating critics to bias the training to the distribution yielded faster discovery of the free-CS constrained to the path, compared to training the model with only one critic. This is evident from \figref{fig:comparisonPathsSampling}, where the partitioned critics captured the local properties of the CS in greater detail.

\begin{figure}[tb]
\centering
\begin{subfigure}{0.49\columnwidth}
	\def\svgwidth{\textwidth}
    %
    %\import{#1.tex}
    %\def\svgwidth{\columnwidth}
    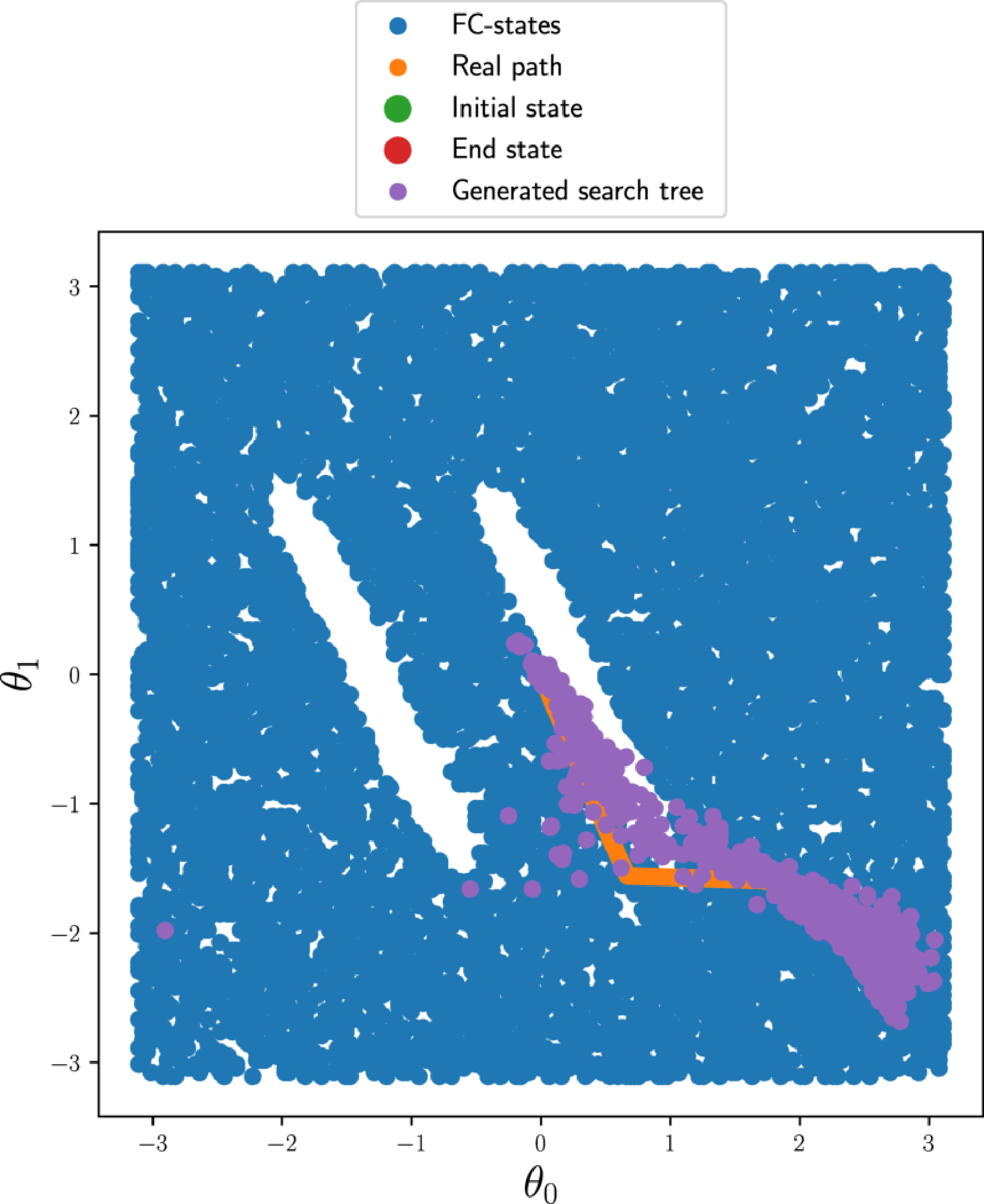

    \caption{Sampling using the model trained  with only the critic for the path.}

    \label{fig:samplingExtrapolationPath}
\end{subfigure}
\hfill
\begin{subfigure}{0.49\columnwidth}
	\def\svgwidth{\textwidth}
    %
    %\import{#1.tex}
    %\def\svgwidth{\columnwidth}
    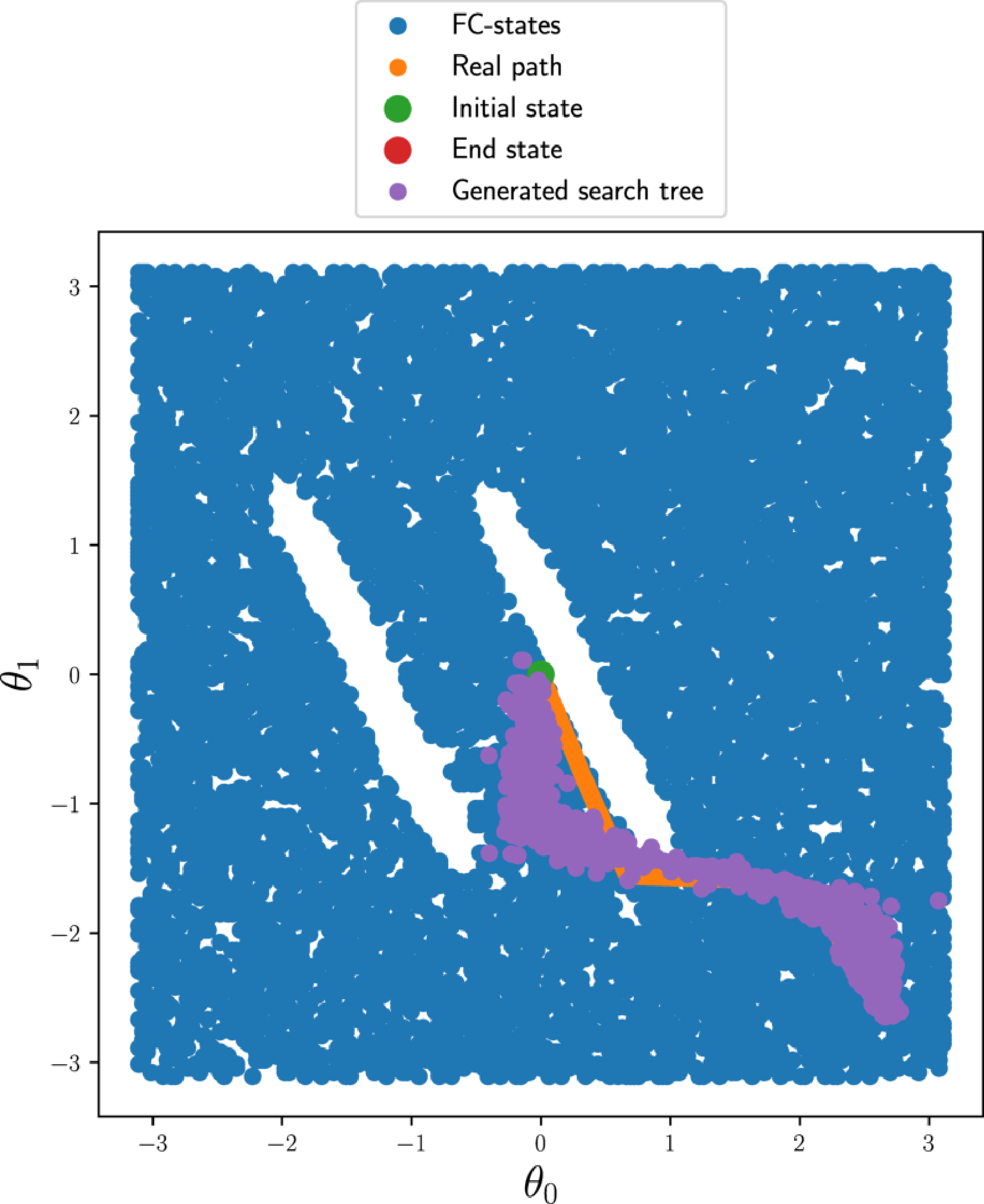

    \caption{Sampling using the model trained  with the critics used to learn the CS and path.}

    \label{fig:samplingExtrapolationPathMultiple}
\end{subfigure}
\caption{Using the GAN model to estimate samples from the path in non-previously seen scenario and CS. The extra critics help to capture information from the whole CS.}
\label{fig:comparisonPathsSamplingExtrapolation}
\end{figure}

Furthermore, incorporating the extra critics helped to provide more information about the complete CS, making it easier to differentiate between CF-states with the image-scenario input, especially in non-previously seen instances. This is exemplified in \figref{fig:comparisonPathsSamplingExtrapolation}, where our model with two critics was able to accurately reconstruct the path compared to the model with only one critic.

\begin{figure}[htb]
    \begin{center}
        \includegraphics[width=.5\textwidth]{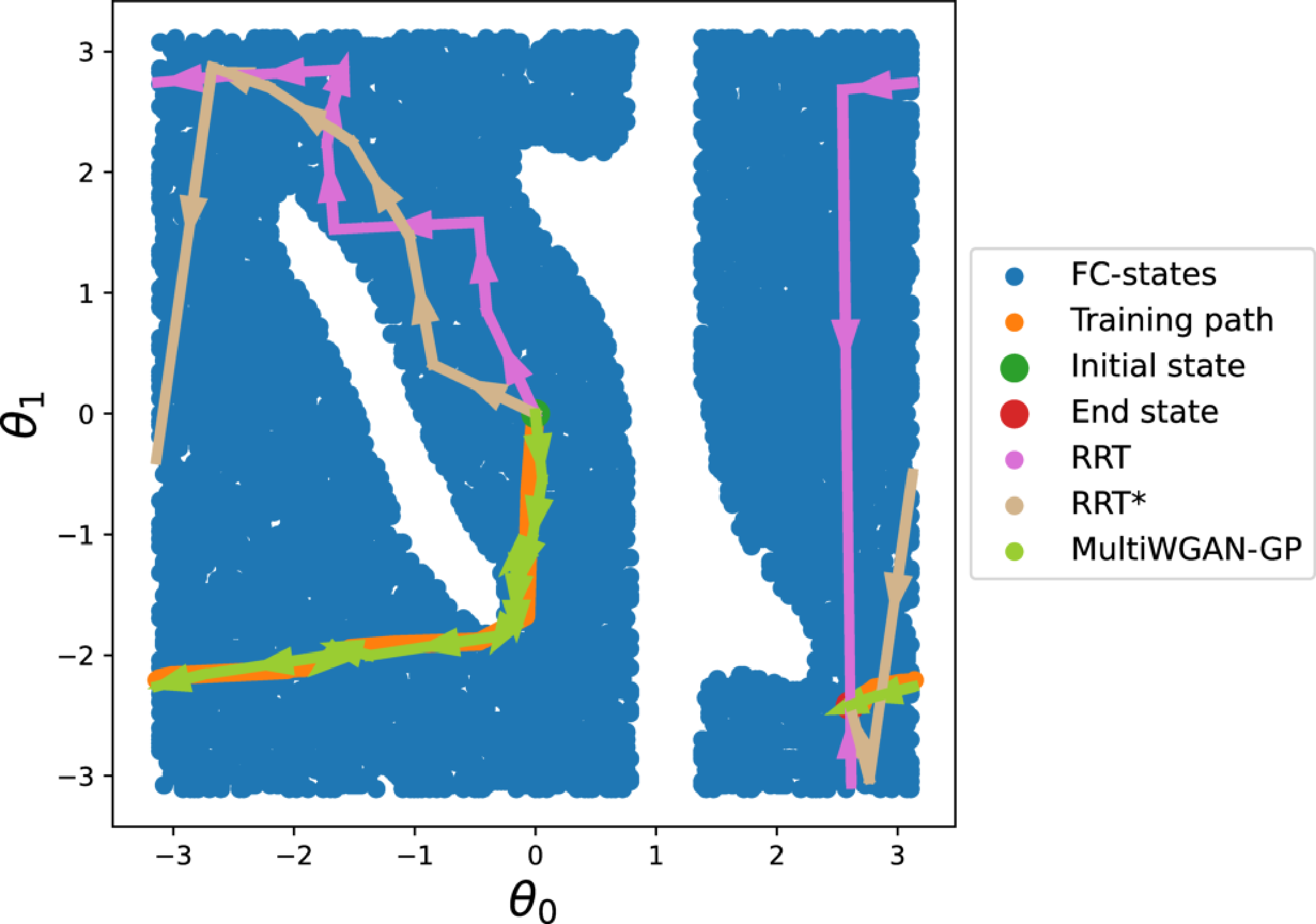}
    \end{center}

\caption{Comparison between RRT, RRT* and the fit data of our method. The arrows indicate the direction of the path.}
    \label{fig:pathFit}
\end{figure}

To implement the trained sampler, we utilized the Open Motion Planning Library (OMPL) \cite{kingston2019exploring} implementations of RRT and RRT*. Specifically, for our proposed MultiWGAN-GP path planner, we replaced the standard uniform sampler with our trained sampler to steer the planner towards the next CF-state. This resulted in a more efficient and effective path planner, as demonstrated in \figref{fig:pathFit}.

\begin{table*}[htbp]
    \tabcolsep=0.09cm
    \footnotesize
\begin{center}
        \begin{tabular}{|c|c|c|c|c|c|}
\toprule
\textbf{Experiment \#} &\textbf{Image-Scenario Input} &\textbf{Algorithm} &\textbf{Path length mean (radians)}&\textbf{Running time mean (seconds)}&\textbf{Success rate}\\\midrule
\multirow{3}{*}{1} & \multirow{3}{*}{\includegraphics[scale=0.033]{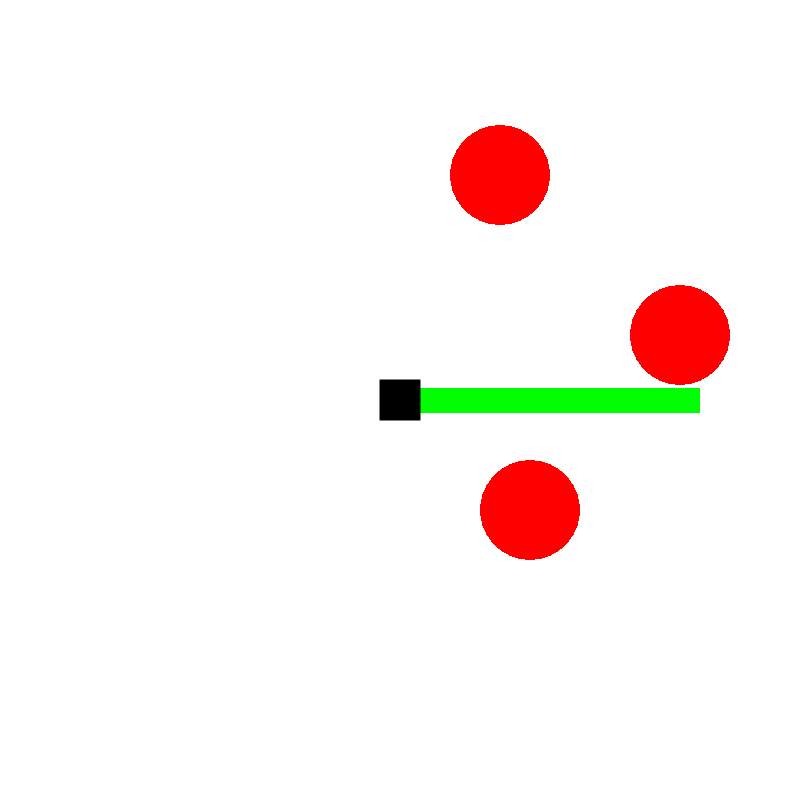}}&RRT&6.45&0.08&70.00\%\\
&&RRT*&4.05&1&70.00\%\\
&&MultiWGAN-GP&4.42&0.21&100.00\%\\\hline
\multirow{3}{*}{2} & \multirow{3}{*}{\includegraphics[scale=0.033]{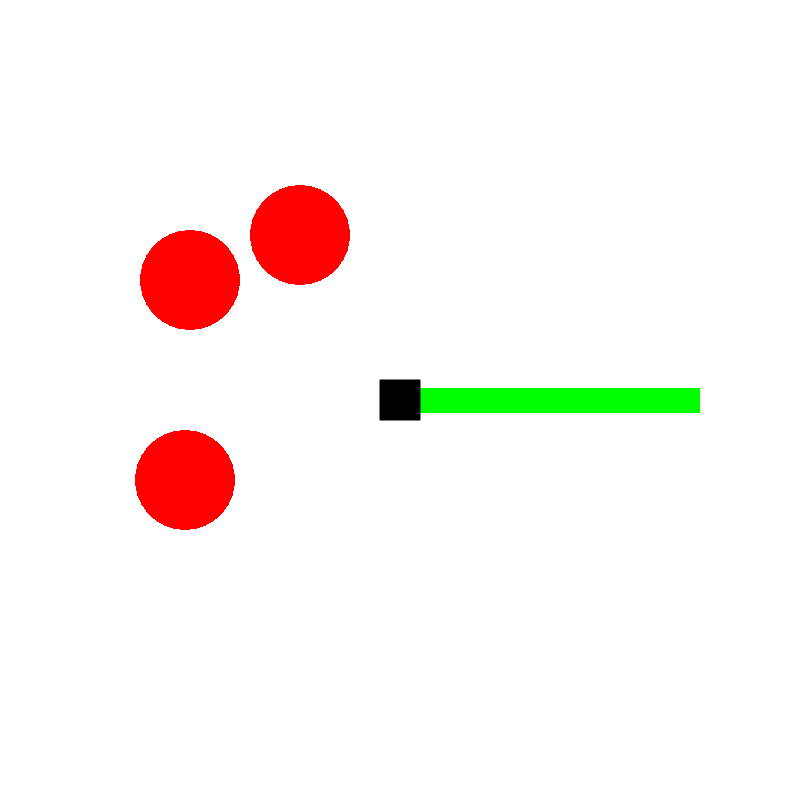}}&RRT&13.74&0.2&70.00\%\\
&&RRT*&14.58&1&50.00\%\\
&&MultiWGAN-GP&15.52&0.32&100.00\%\\\hline
\multirow{3}{*}{3} & \multirow{3}{*}{\includegraphics[scale=0.033]{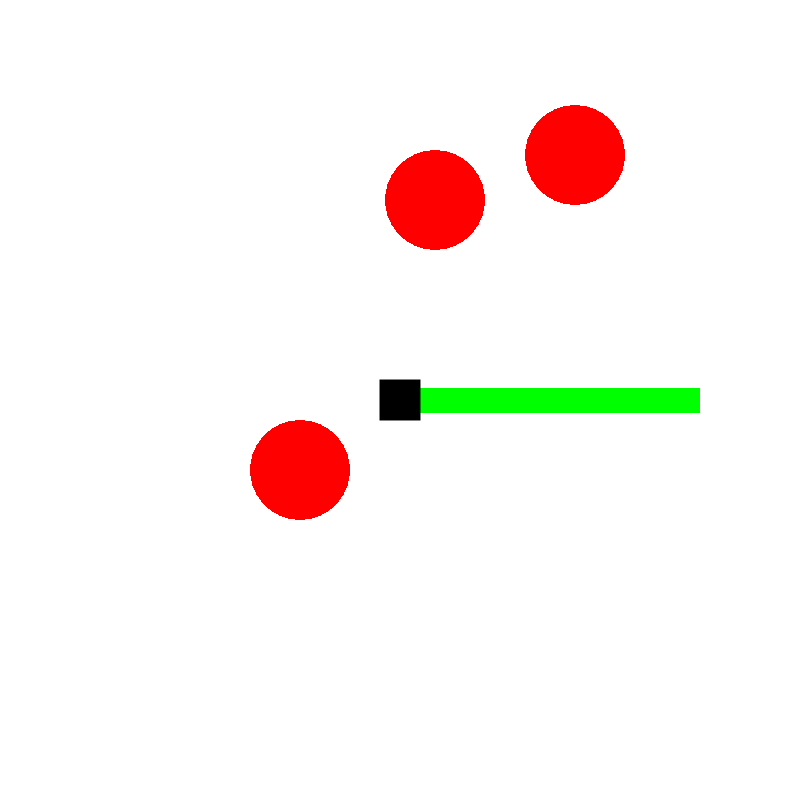}}&RRT&14.32&0.27&70.00\%\\
&&RRT*&12.82&1&30.00\%\\
&&MultiWGAN-GP&10.04&0.22&100.00\%\\\hline
\multirow{3}{*}{4} & \multirow{3}{*}{\includegraphics[scale=0.033]{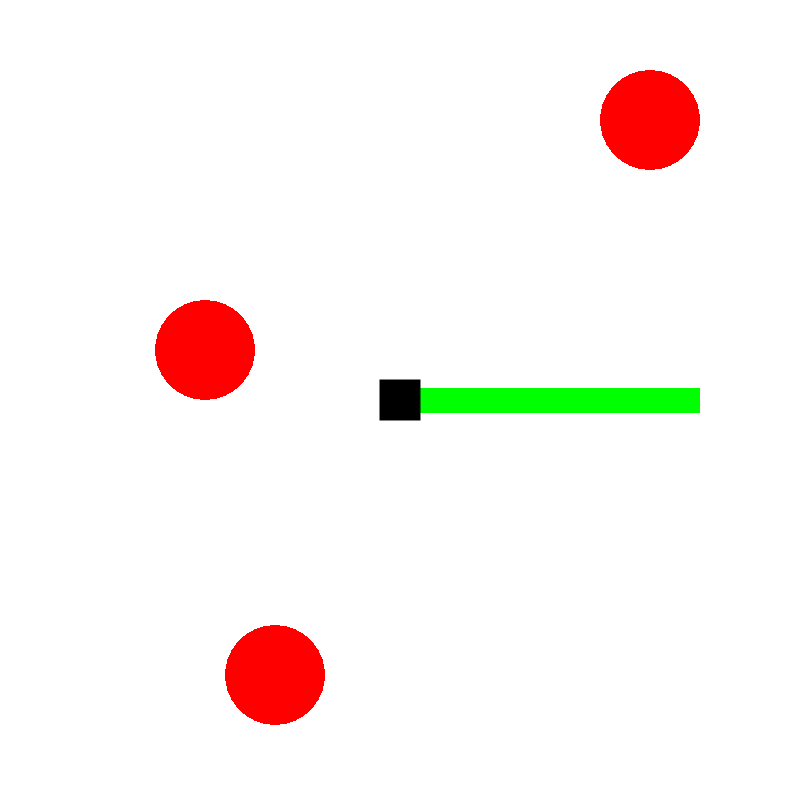}}&RRT&5.85&0.18&80.00\%\\
&&RRT*&5.57&1&60.00\%\\
&&MultiWGAN-GP&5.79&0.18&100.00\%\\\hline
\multirow{3}{*}{5} & \multirow{3}{*}{\includegraphics[scale=0.033]{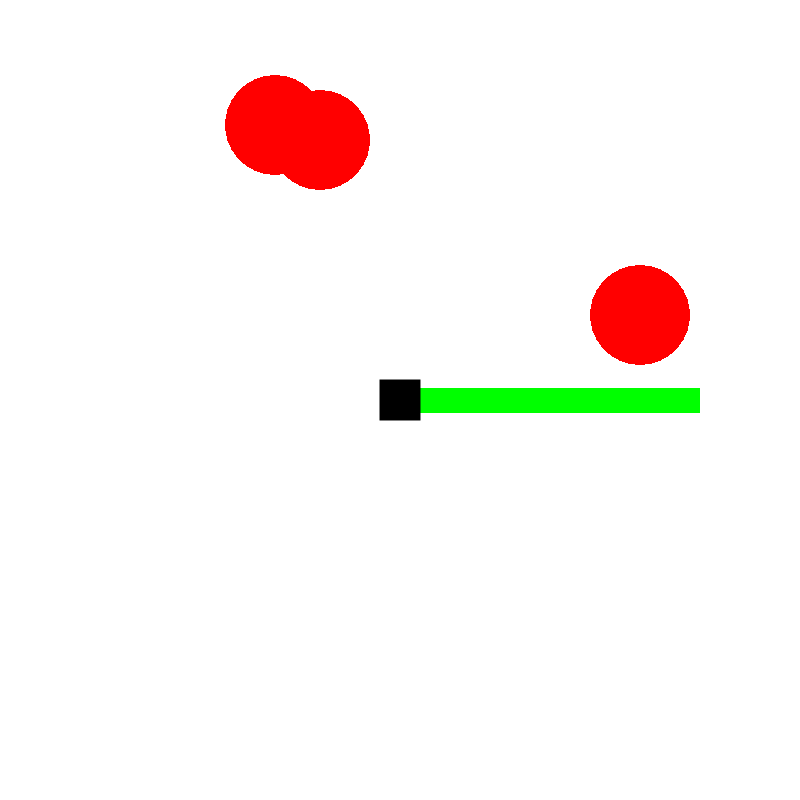}}&RRT&8.83&0.13&80.00\%\\
&&RRT*&6.59&1&70.00\%\\
&&MultiWGAN-GP&7.54&0.17&100.00\%\\\hline
\multirow{3}{*}{6} & \multirow{3}{*}{\includegraphics[scale=0.033]{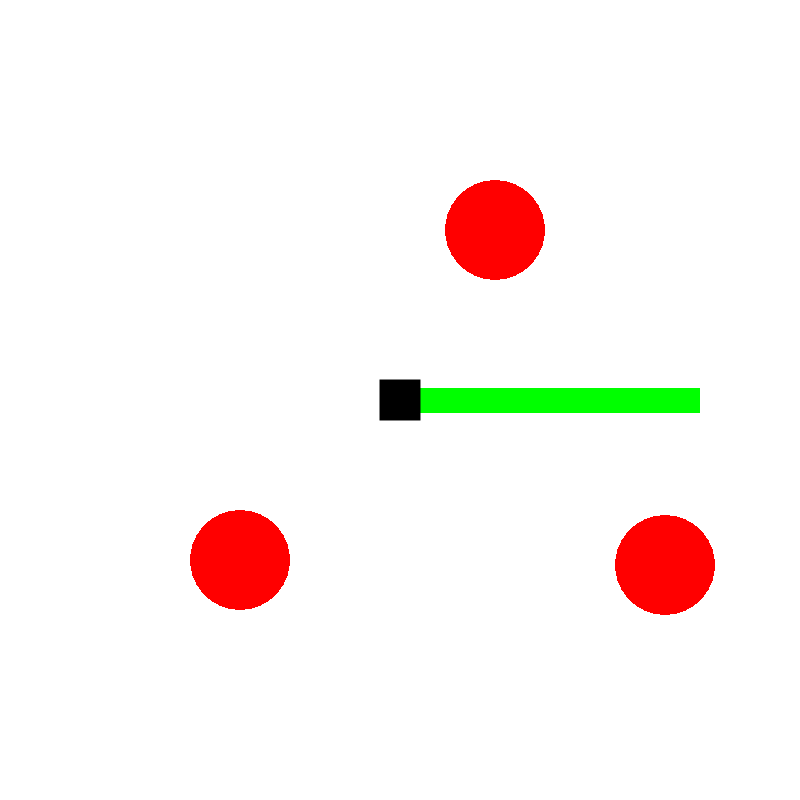}}&RRT&15.12&0.2&70.00\%\\
&&RRT*&14.68&1&80.00\%\\
&&MultiWGAN-GP&18.11&0.23&100.00\%\\\hline
\multirow{3}{*}{7} & \multirow{3}{*}{\includegraphics[scale=0.033]{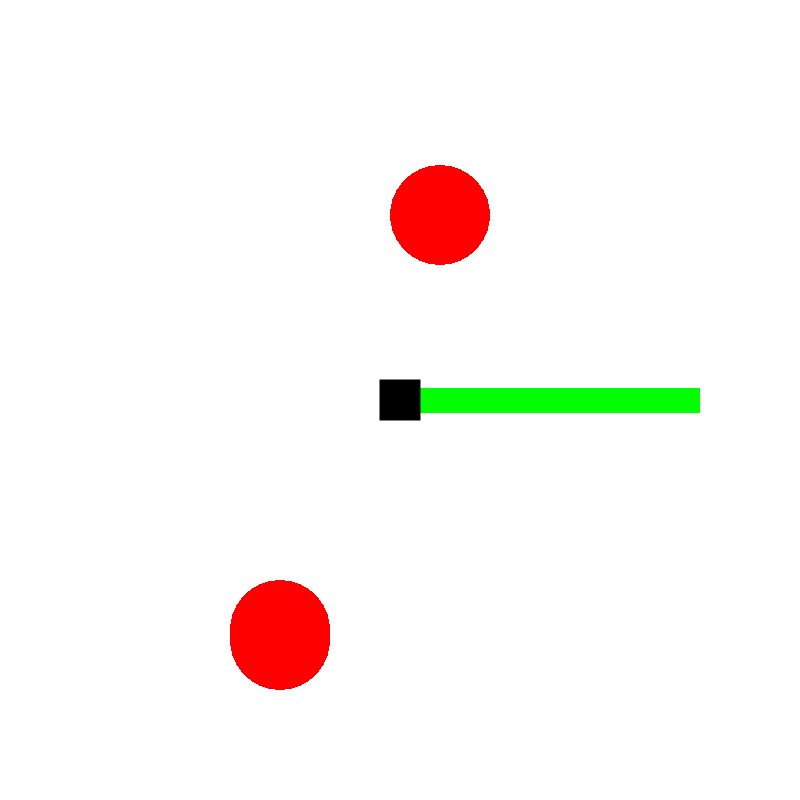}}&RRT&15.68&0.18&80.00\%\\
&&RRT*&15.98&1&80.00\%\\
&&MultiWGAN-GP&17.87&0.25&100.00\%\\\hline
\multirow{3}{*}{8} & \multirow{3}{*}{\includegraphics[scale=0.033]{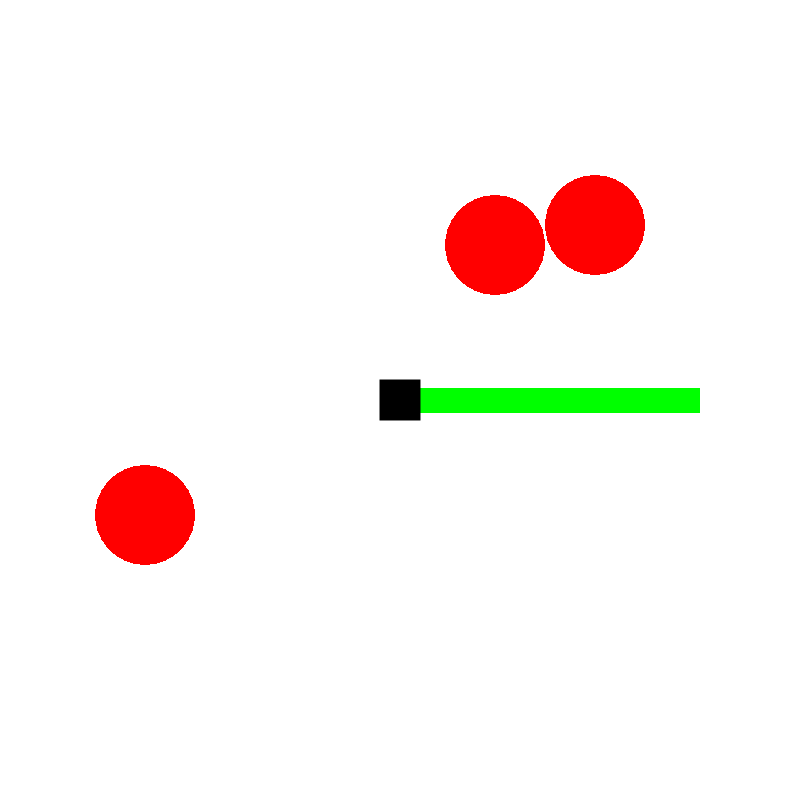}}&RRT&14.5&0.2&70.00\%\\
&&RRT*&15.08&1&70.00\%\\
&&MultiWGAN-GP&16.56&0.21&100.00\%\\\hline
\multirow{3}{*}{9} & \multirow{3}{*}{\includegraphics[scale=0.033]{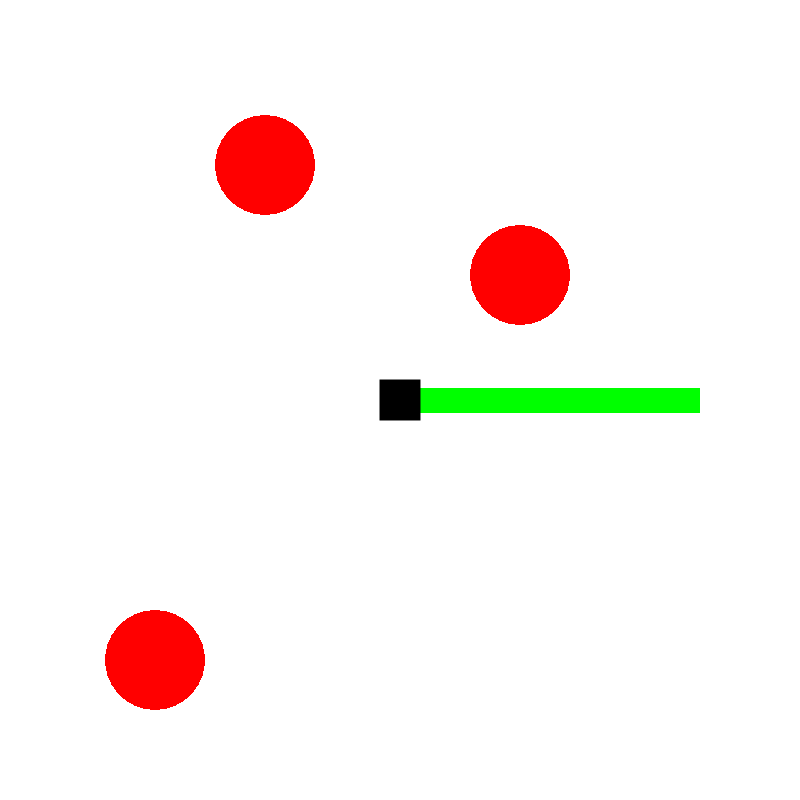}}&RRT&11.35&0.13&80.00\%\\
&&RRT*&11.2&1&70.00\%\\
&&MultiWGAN-GP&10.27&0.17&100.00\%\\\hline
\multirow{3}{*}{10} & \multirow{3}{*}{\includegraphics[scale=0.033]{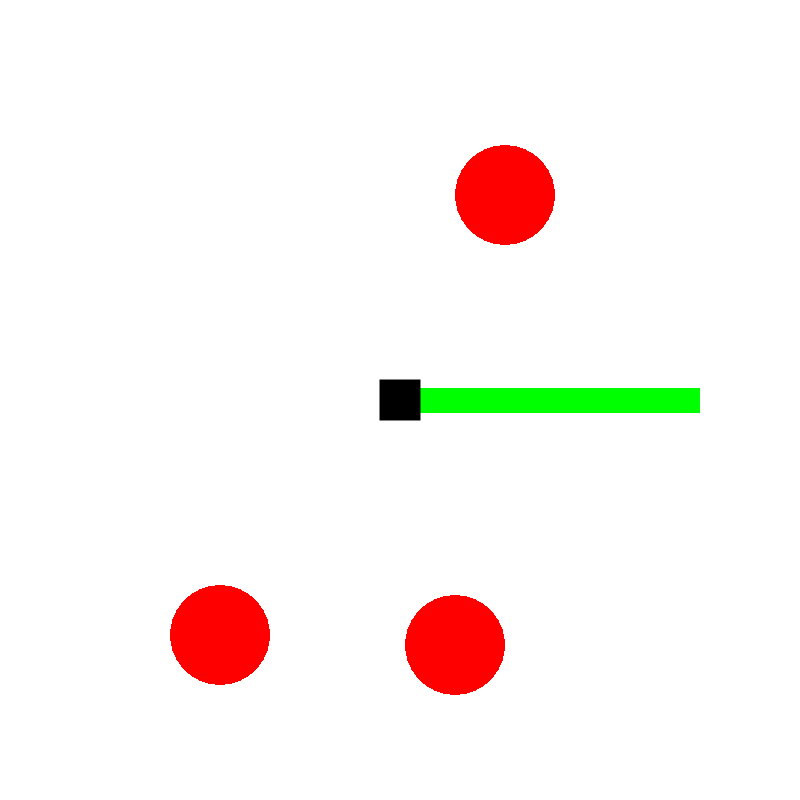}}&RRT&10.43&0.14&20.00\%\\
&&RRT*&5.81&1&80.00\%\\
&&MultiWGAN-GP&4.41&0.16&100.00\%\\
            \bottomrule
\end{tabular}
\end{center}
    \caption{The results of the path planning task for extrapolation indicate that our method successfully improved the success rate of RRT without a significant increase in running time.}
    \label{tab:table1}
\end{table*}

We conducted experiments to directly compare our method to RRT and RRT* on 10 previously unseen scenarios. We aimed to demonstrate improvements in both the quality of the path, represented by the path length  mean of RRT* , and the running mean time. Each of the 10 different scenarios were run 100 times to obtain the estimated scores. As the CS is relatively simple, we allowed only 1 second to solve each of the paths. The initial state of the path and the final state goal were fixed for all the CSs during these tests. The results of these experiments is given in Table \ref{tab:table1}.

\begin{figure}[tb]
\centering
\begin{subfigure}{0.49\columnwidth}
	\def\svgwidth{\textwidth}
    %
    %\import{#1.tex}
    %\def\svgwidth{\columnwidth}
    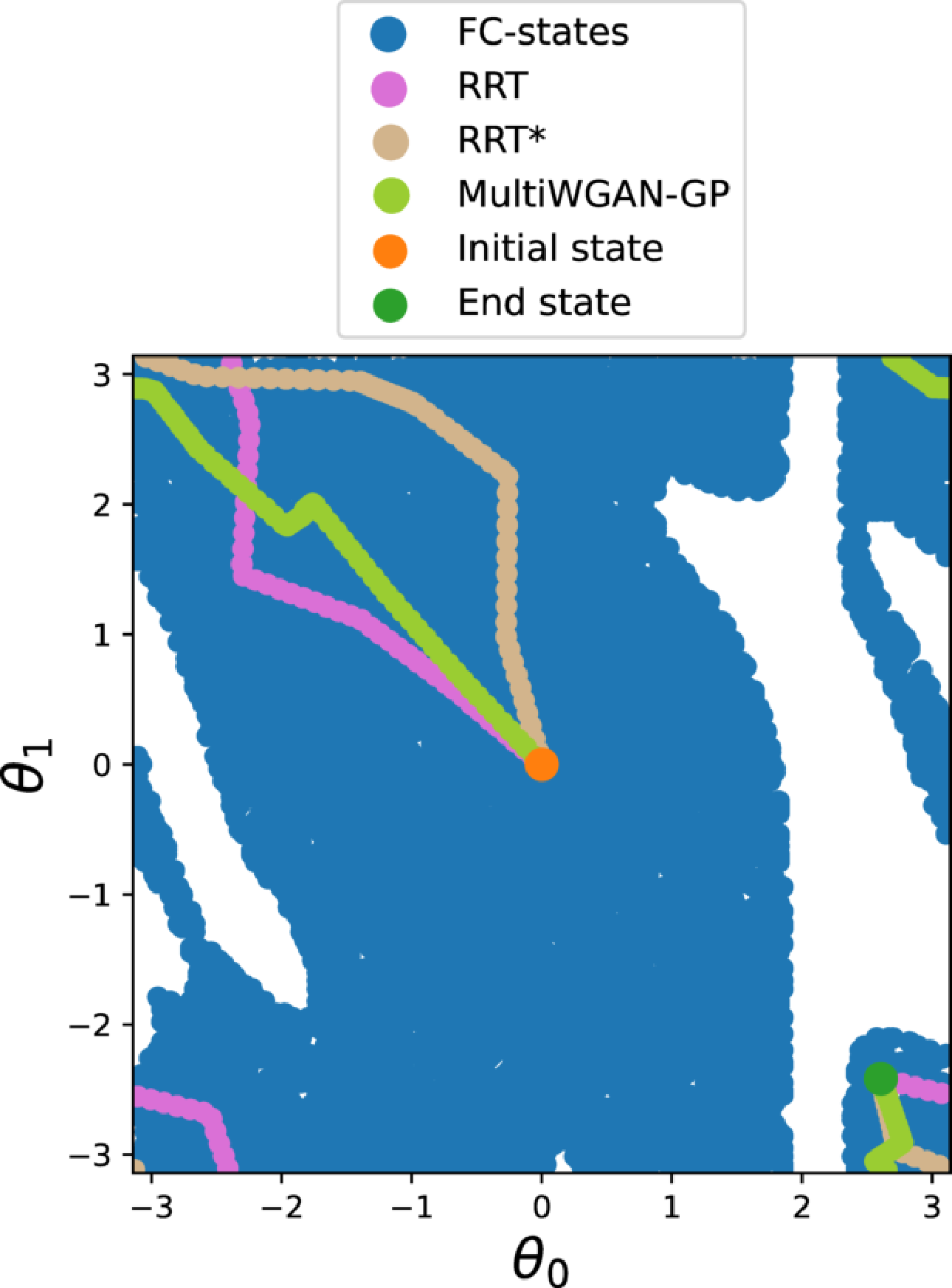

    \caption{Our method produced longest path in this case. There is no sample in the training data that are close to this unseen scenario.}

\end{subfigure}
\hfill
\begin{subfigure}{0.49\columnwidth}
	\def\svgwidth{\textwidth}
    %
    %\import{#1.tex}
    %\def\svgwidth{\columnwidth}
    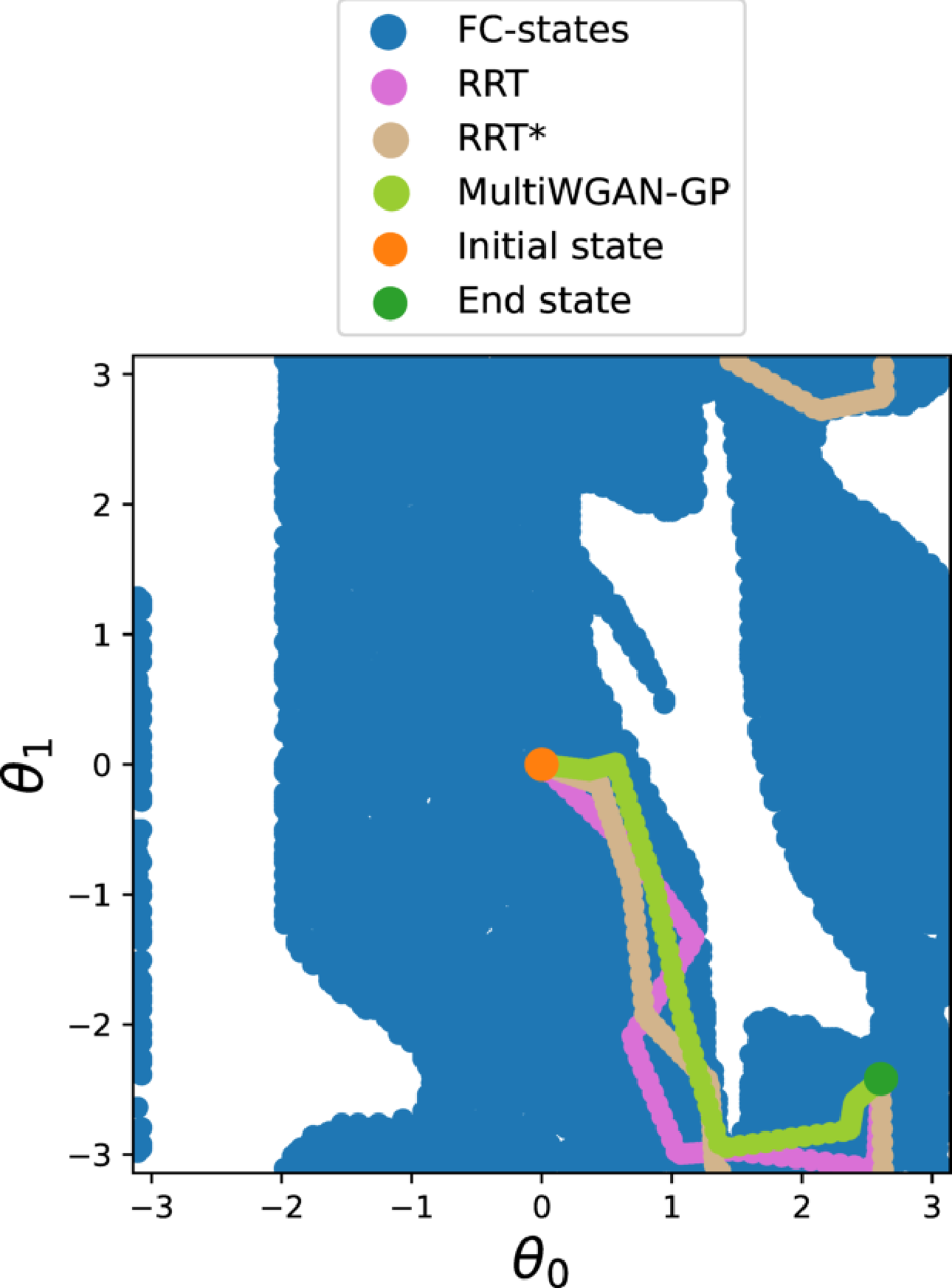

    \caption{Our method produces the shortest path. There are some samples of the training dataset that are similar to this unseen scenario.}

\end{subfigure}
\caption{Extrapolation of the planners on non-previously seen scenarios. Our method is able to find the shortest path in the scenarios that are similar to the trained data.}
\label{fig:pathsExtrapolations1}
\end{figure}

The results reported in Table \ref{tab:table1} indicate that there were instances where the shortest path could not be found by the algorithms. However, our proposed method, MultiWGAN-GP, was successful in shortening the path in 30\% of cases, while RRT* and RRT achieved reductions of 40\% and 30\%, respectively. It is worth noting that these cases typically occurred when the CS and scenario were significantly different from the training data. This is because the learned model produces a latent vector close to several similar scenarios, some of which may not be precisely useful for the current scenario to generate the shortest path, as shown in \figref{fig:pathsExtrapolations1}. Furthermore, the implementation of RRT does not utilize rewiring, which could potentially improve the path length with the remaining planning time left for the planner to find the CF-path.

The experimental results demonstrate that our method can generate a feasible CF-path for the given scenario. With more data, it may be able to achieve better path length than RRT* when the planning time is constrained.

\begin{figure}
\begin{center}
\includegraphics[width=.45\textwidth]{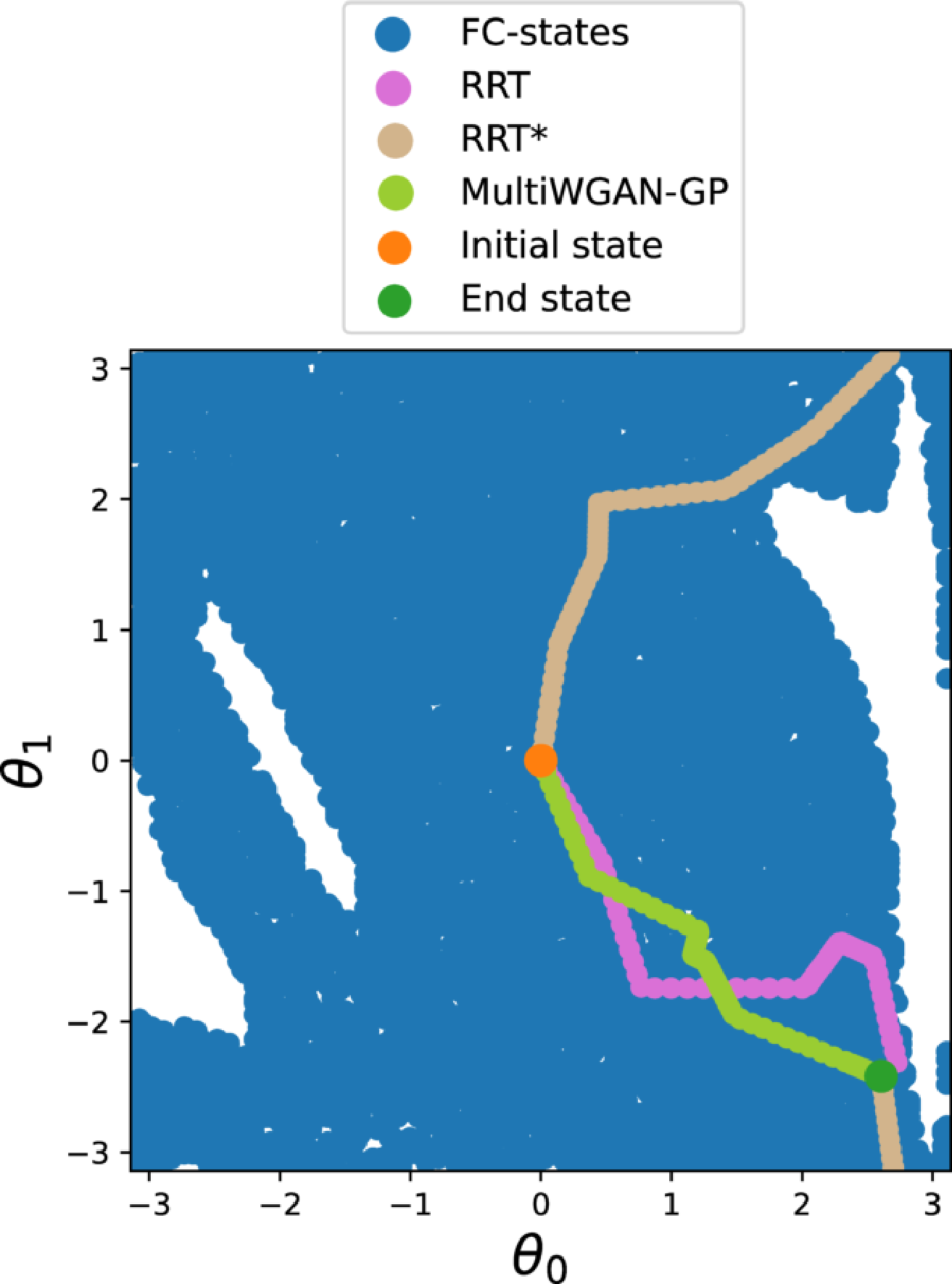}
\caption{Extrapolation of the path where our method achieves the same speed as the RRT algorithm, taking into account the overhead of calling both the encoder and generator.}
\label{fig:experiment4extrapolation}
\end{center}

\end{figure}

Our proposed algorithm was able to find a CF-path in all the new cases in a timely manner compared with the running time of RRT. During testing, we found that we needed to sample at least 300 states from the generated path, which takes approximately 0.18 seconds, one example is presented in \figref{fig:experiment4extrapolation}. Since we used path interpolation as our training data, most of the states generated by the neural network are in close proximity to each other, which slows down the steering of the algorithm towards the closest neighbor in the graph.

To speed up our method, we need to reduce the number of sampling points and ensure that each point is as close as possible to a waypoint that can be accepted by the algorithm. This can be achieved by increasing the amount of training paths to cover a variety of scenarios, allowing the model to learn the minimum number of waypoints needed to reduce the amount of required collision checking during interpolation between the new state and the current state. By improving the model's ability to learn these minimal waypoints, we can reduce the time taken to find a path and increase the efficiency of our proposed method.

There is still room for improvement in terms of the speed of querying the encoder and generator to achieve real-time performance. One approach to achieve this could be reducing the number of operations required by the neural network, for instance, by optimizing the architecture or using more efficient algorithms.

Overall, while our proposed method shows promising results, there is still further work to be done to improve its efficiency and real-time performance. By incorporating the aforementioned improvements, we believe that our algorithm can be further optimized to achieve better results and be applicable to more complex scenarios.

Regarding the success rate, our algorithm was able to generate feasible CF paths each time, even when faced with previously unknown scenarios, whereas RRT and RRT* struggled to find CF waypoints. This is especially important in tasks where low latency and safety are major concerns, such as human-robot interaction. 

Although our planning scenarios were generated randomly, we recognize the possibility of our algorithm encountering challenges in generating a CF path within the time constraints of the query. Furthermore, if the dataset exhibits bias towards certain regions, the probability of sampling far from the mean vector ($\vector{\mu}$) diminishes exponentially, even with a large variance that approximates a uniform distribution. In such cases, it may be necessary to acquire additional training data that encompasses a broader range of scenarios to address these situations effectively.

\subsection{Discussion}

In this study, we have discovered that implementing VAEs and WGAN-GP for CS reconstruction within multimodal models, specifically Image-to-CS, poses a significant challenge. The complexity of the model must be minimized by reducing the number of operations the neural network needs to learn the topology based on the image. By simplifying the model, we were able to train the WGAN-GP and VAE independently, resulting in improved training times for both models.

In addition, the study showed that topology metrics can be used to evaluate the ability of GANs to represent a distribution. However, we found that this approach requires the transformation of data to accommodate for other structures that may be present in CSs used in robotic tasks, in addition to holes. By transforming the data in such a way, other structures can be described using the rank of topology. This approach can be beneficial in assessing the quality of GAN-generated CSs to be used in robotic planning and control applications.

The experiments pointed out that using local critics in WGAN models can improve their performance, even in different tasks such as path planning where the reconstruction of the configuration space and the generation of collision-free paths are related problems but not identical. The inclusion of local critics helped the model to converge to a better distribution and improve its ability to generate valid paths. However, it is worth noting that the use of multiple critics can lead to situations where one partition of the data does not generate any points, which can limit the effectiveness of the training process for that critic.

We have demonstrated that the integration of VAEs and WGAN-GP is an effective approach to accelerate path planning in 4-dimensional CSs, resulting in higher percentage of CF-paths in previously unseen scenarios and quasi-optimal paths when the WSs are similar to the training data. However, it is important to note that minimizing the number of model queries is crucial for real-time path planning applications.

\section{Conclusion and Future Work}

In this work, we have presented a novel approach for training WGAN-GP models conditioned by VAEs, which can be utilized for tasks related to CS reconstruction and path planning. We also proposed to use homology ranking to measure the reconstruction of the configuration space by using the complement of the reconstructed data to be able to measure the clusters of CF-states as holes in the complement of the configuration space and being able to discriminate which models describe better the reconstruction. In addition, we explored the use of local critics to improve the reconstruction of the CF-states in the CS for path planning tasks.

 The results of our experiments demonstrate that our proposed model is capable of generating CF-paths in unknown scenarios with improved success rate and reduced running time when compared to conventional path planning algorithms such as RRT and RRT*.

We have also demonstrated the effectiveness of our proposed method in planning paths in a 4-dimensional space for a  2-dimensional 2-DOF simulated robot. However, to establish the broader applicability of our method, it is necessary to extend it to higher dimensional spaces for redundant articulated robots. This is critical to show its usefulness in solving high-dimensional problems for real-world applications and using real robots. Future work will  focus on exploring the feasibility of this extension and evaluating the performance of our method on more complex tasks and scenarios.

One of the challenges that we identified is the processing of the input data. In our work, we used a 2D representation of the configuration space for the robot. However, in real-world scenarios, we need to combine the VAE with noisy depth information to provide sufficient information to the GAN model for reconstructing the configuration space of the robot. Therefore, further research is needed to explore methods for incorporating depth information into the VAE-GAN framework to improve the accuracy and robustness of the model.

\section*{Acknowledgement}

Jorge Ocampo-Jimenez is a doctorate student funded by Consejo Nacional de Ciencia y Tecnología (CONACyT, Mexico City, Grant No. 278823) and by the merit scholarship programme for foreign students (PBEEE) from the Fonds de recherche du Québec – Nature et technologies. This work was also partly supported by the Natural Sciences and Engineering Research Council of Canada (NSERC). Additionally, we acknowledge the valuable assistance and resources provided by Calcul Qu\'ebec and the Digital Research Alliance of Canada.

\bibliographystyle{IEEEtran}

\bibliography{IEEEabrv,bibtex/bib/refs} 

\begin{IEEEbiography}[{\includegraphics[width=1in,height=1.25in,clip,keepaspectratio]{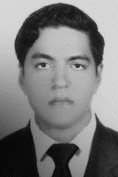}}]{Jorge Ocampo-Jimenez} received his M.Sc. degree in Computational Sciences from the National Institute for Astrophysics, Optics and Electronics (INAOE), Mexico, in 2010. He worked as a professional researcher at INAOE, from 2011 to 2017. He is currently a Ph.D. student at the University of Sherbrooke, Canada. His research interests include collaborative robotics, machine learning and motion planning.
\end{IEEEbiography}
\vskip 0pt plus -1fil
\begin{IEEEbiography}[{\includegraphics[width=1in,height=1.25in,clip,keepaspectratio]{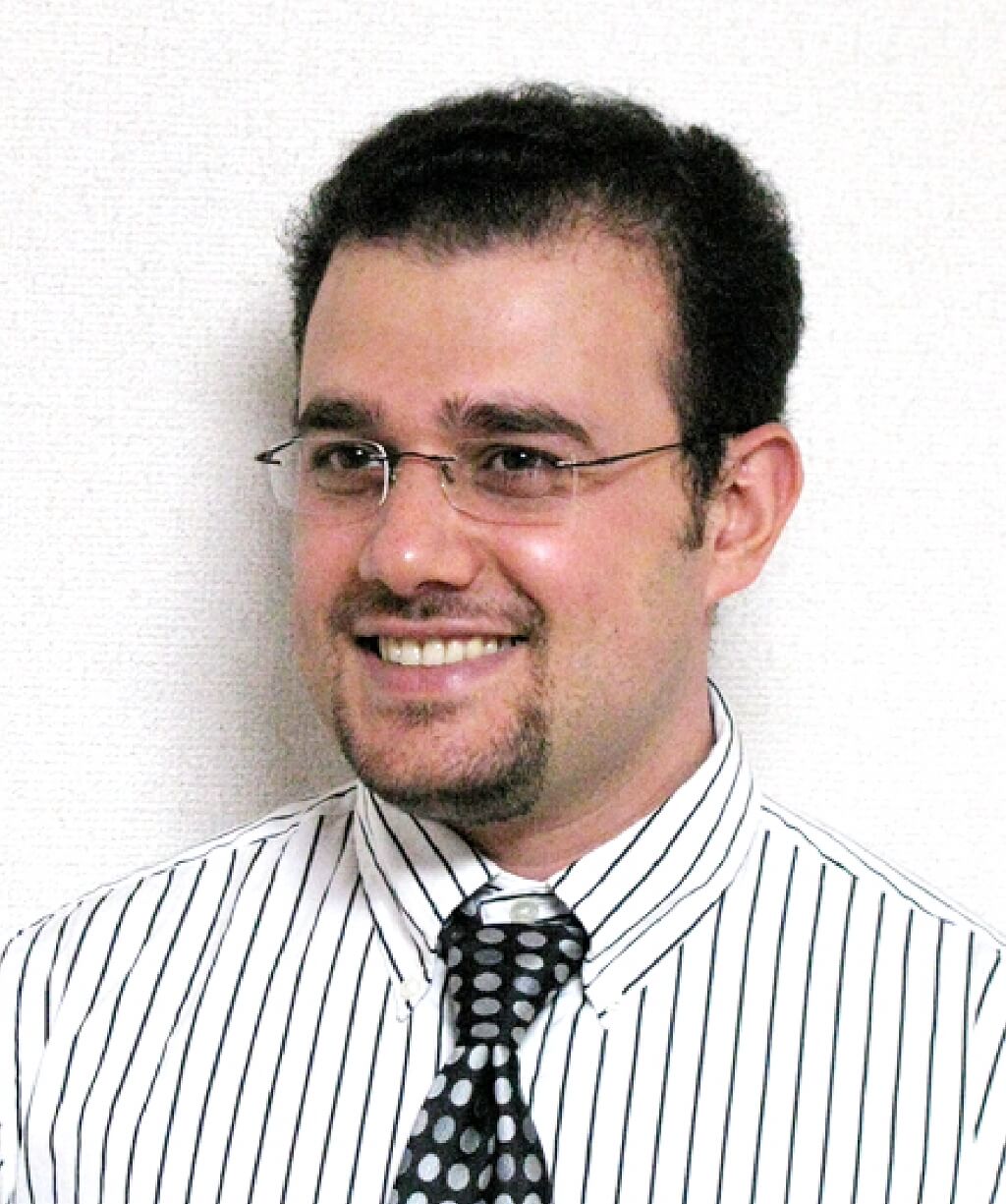}}]{Wael Suleiman} received the Master's and Ph.D. degrees in automatic control from Paul Sabatier University, Toulouse, France in 2004 and 2008, respectively. He has been Postdoctoral researcher at AIST, Tsukuba, Japan from 2008 to 2010, and at Heidelberg University, Germany from 2010 to 2011. He joined University of Sherbrooke, Quebec, Canada, in 2011, and is currently Full Professor at Electrical and Computer Engineering Department. His research interests include collaborative and humanoid robots, motion planning, nonlinear system identification and control and numerical optimization. 
\end{IEEEbiography}

\appendices

  \section{Neural Network architecture}

In this section, we discuss each component of the proposed architecture in detail.

\begin{figure*}[ht]
\begin{center}
\includegraphics[width=\textwidth]{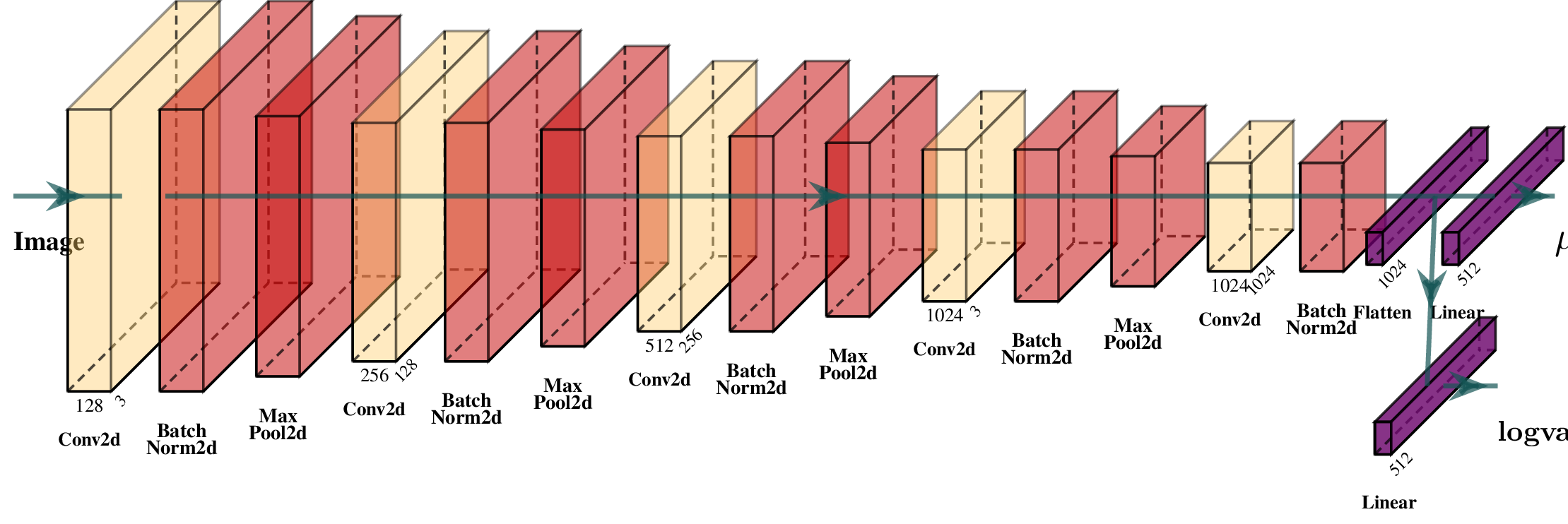}
\caption{Encoder network.}
\label{fig:encoder}
\end{center}
\end{figure*}

First, let us consider the encoder. To design the encoder, we partitioned the image into four small batches. This approach not only allows for faster processing of the latent vector but also helps capture more local information when multiple obstacles appear in the same region. The choice of convolutions in the encoder follows a similar architecture to the one used in \cite{park2019SPADE} for VAE-GAN models. However, we reduced the number of convolutional layers to decrease the model size and improve query speed. The architecture of the encoder is given in \figref{fig:encoder}.

\begin{figure*}[ht]
\centering
\includegraphics[width=\textwidth]{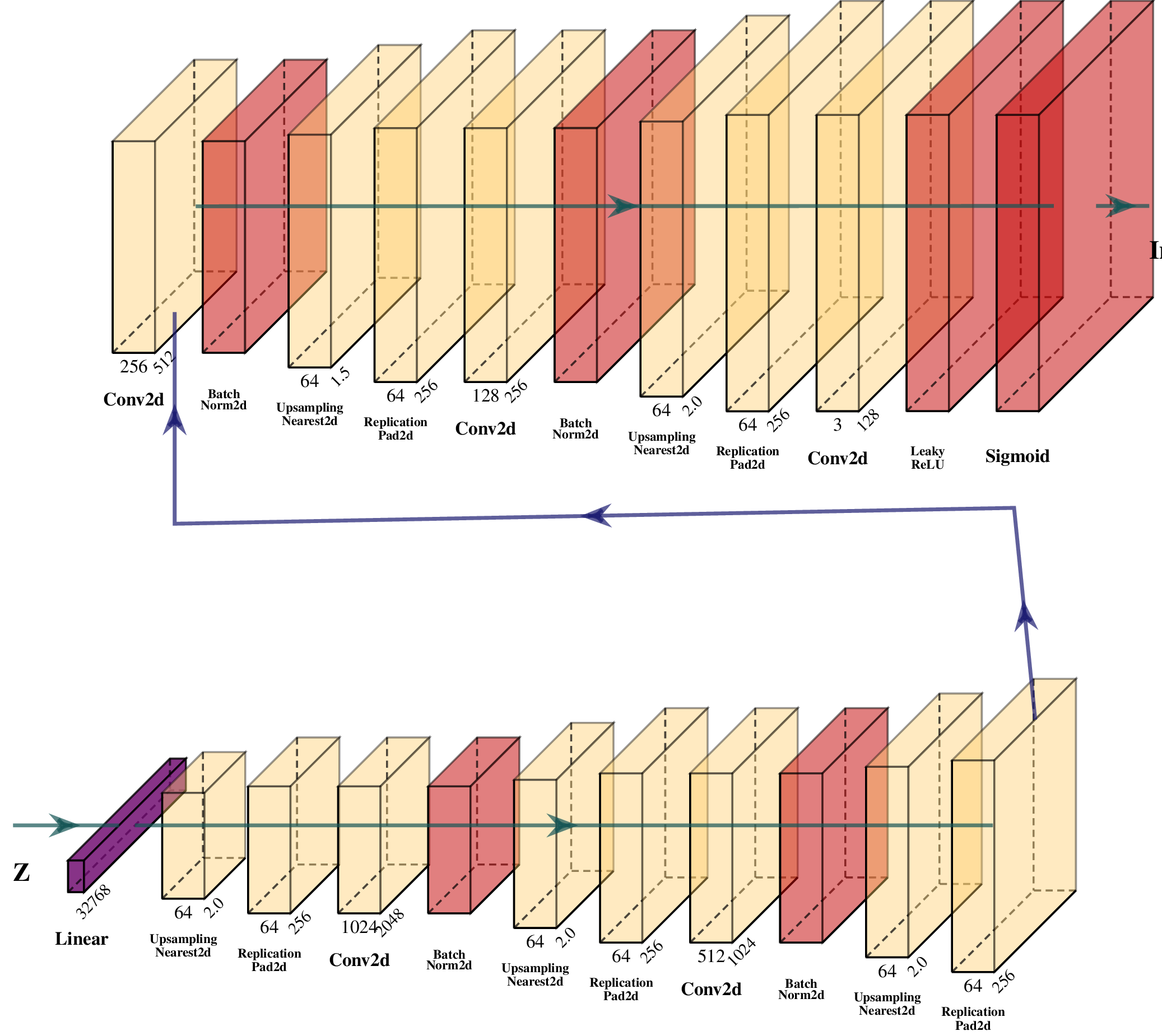}
    \caption{Decoder network.}
    \label{fig:decoder}
\end{figure*}

Moving on to the decoder, as depicted in \figref{fig:decoder}, we employed deconvolution techniques based on \cite{odena2016deconvolution} to reduce artifacts during image reconstruction in VAE training.

\begin{figure*}[ht]
\centering
\includegraphics[width=.7\textwidth]{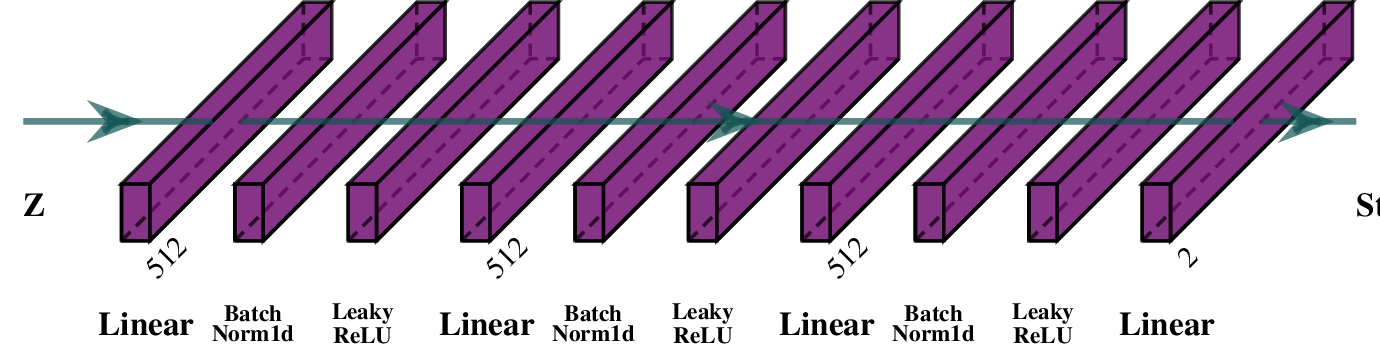}
    \caption{Generator network.}
    \label{fig:generator}
\end{figure*}

For the generator, shown in \figref{fig:generator}, we adopted the architecture proposed in \cite{10.5555/3295222.3295327} to sample a $\mathbb{R}^n$ vector from the latent variable $\vector{z}$.

\begin{figure*}[ht]
\centering
\includegraphics[width=.7\textwidth]{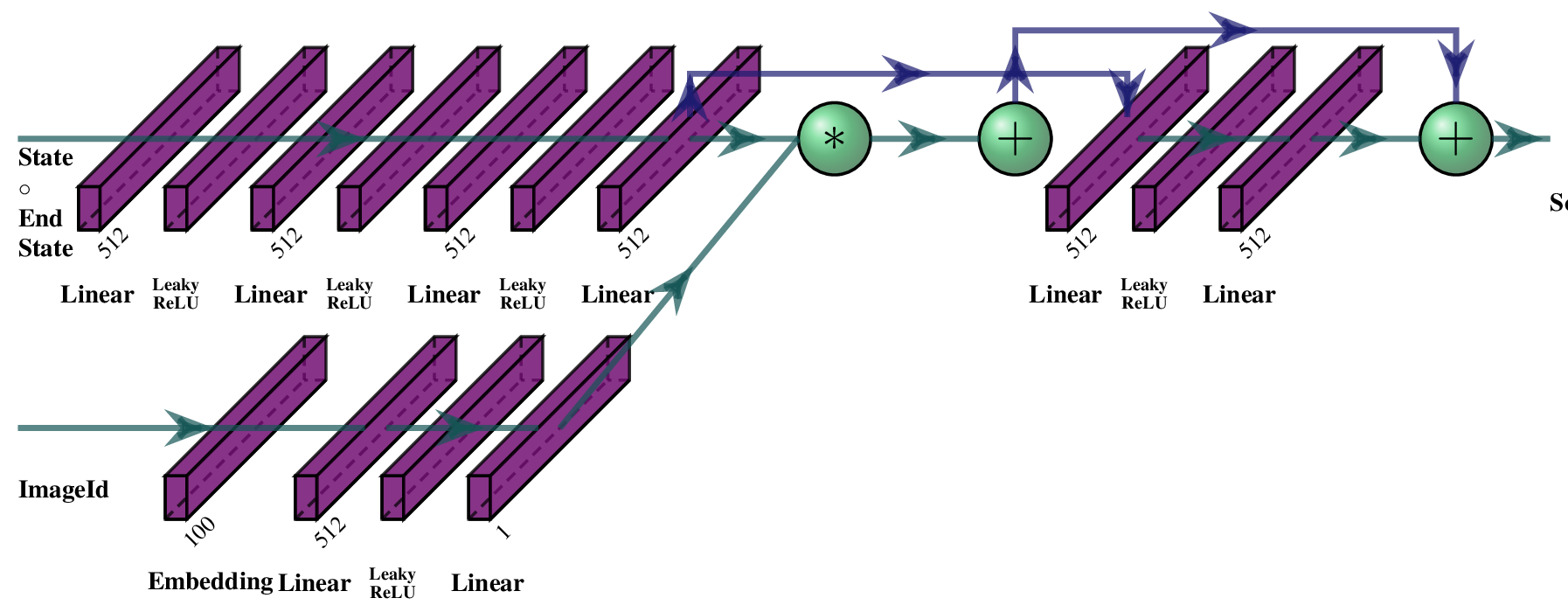}
    \caption{Critic network.}
    \label{fig:critic}
\end{figure*}

Lastly, let us consider the critic. In contrast to image-image generation models, our model employs a unique identifier to represent the image instead of processing it directly. The purpose of this unique ID is to indicate which instance of the current configuration space we are dealing with during training, rather than evaluating the quality of the reconstructed scenario. Therefore, a significant portion of the model is dedicated to distinguishing between the generated vectors representing collision-free states. We adapted the proposal from \cite{DBLP:conf/iclr/MiyatoK18} to project the condition onto the critic, thereby improving the performance of the trained generator. \figref{fig:critic} provides an illustration of the critic component.

%\begin{figure*}[ht]
%\begin{center}
%\include{encoder}
%\caption{Encoder network.}
%\label{fig:encoder}
%\end{center}
%\end{figure*}
%\begin{figure*}[ht]
%\centering
%\include{generator}
%    \caption{Generator network.}
%    \label{fig:generator}
%\end{figure*}
%\begin{figure*}[ht]
%\centering
%\include{critic}
%    \caption{Critic network.}
%    \label{fig:critic}
%\end{figure*}
%
%\begin{figure*}[ht]
%\centering
%\include{decoder}
%    \caption{Decoder network.}
%    \label{fig:decoder}
%\end{figure*}


  \label{FirstAppendix}

\end{document}